\documentclass{article}

\usepackage{microtype}
\usepackage{graphicx}
\usepackage{subfigure}
\usepackage{booktabs} 
\usepackage{spverbatim}
\usepackage{hyperref}

\usepackage[accepted]{icml2025}

\usepackage{amsmath}
\usepackage{amssymb}
\usepackage{mathtools}
\usepackage{amsthm}
\usepackage{titletoc}
\usepackage{enumitem}

\usepackage[capitalize,noabbrev]{cleveref}

\theoremstyle{plain}

\theoremstyle{definition}

\theoremstyle{remark}

\usepackage[textsize=tiny]{todonotes}

\newcommand{\name}{GeoManip}
\newcommand\erran[1]{
\ifthenelse{\equal{\showcomments}{}}{{\color{blue} erran: #1}}{\ignorespaces}
}

\definecolor{citecolor}{HTML}{0071BC}
\hypersetup{colorlinks,linkcolor={red},citecolor={citecolor}}

\title{\hspace{-2pt}\name: Geometric Constraints as General Interfaces \\ for Robot Manipulation}
\author{
\textbf{Weiliang Tang$^{1,2}$, Jia-Hui Pan$^2$, Yun-Hui Liu$^2$, Masayoshi Tomizuka$^1$, Li Erran Li$^3$}, \\ \textbf{Chi-Wing Fu$^2$\thanks{Corresponding authors. This work is partially supported by the InnoHK of the Government of Hong Kong via the Hong Kong Centre for Logistics Robotics and the CUHK T Stone Robotics Institute.}, Mingyu Ding$^{1,4}$}
\vspace{4pt} \\
$^1$UC Berkeley \quad $^2$CUHK \quad $^3$AWS AI, Amazon \quad $^4$UNC-Chapel Hill
\vspace{4pt} \\
\href{http://geoconstraintmanip.github.io}{http://geoconstraintmanip.github.io}
}

\icmltitlerunning{Geometric Constraints as General Interfaces for Robot Manipulation}

\begin{document}
\date{}
\maketitle




\begin{abstract}

%
%
%
We present \name{}, a framework to enable generalist robots to leverage essential conditions derived from object and part relationships, as geometric constraints, for robot manipulation.
For example, cutting the carrot requires adhering to a geometric constraint: the blade of the knife should be perpendicular to the carrot's direction.
By interpreting these constraints through symbolic language representations and translating them into low-level actions, \name{} bridges the gap between natural language and robotic execution, enabling greater generalizability across diverse even unseen tasks, objects, and scenarios.
Unlike vision-language-action models that require extensive training, \name{} operates training-free by utilizing large foundational models: a constraint generator that predicts stage-specific geometric constraints and a geometry parser that identifies object parts involved in these constraints. A solver then optimizes trajectories to satisfy inferred constraints from task descriptions and the scene.
Furthermore, \name{} learns in-context and provides five appealing human-robot interaction features: on-the-fly policy adaptation, learning from human demonstrations, learning from failure cases, long-horizon action planning, and efficient data collection for imitation learning.
Extensive evaluations on both simulations and real-world scenarios demonstrate \name{}’s state-of-the-art performance, with superior out-of-distribution generalization while avoiding costly model training.

\end{abstract}

\begin{figure}
    \centering
    \includegraphics[width=0.99\linewidth]{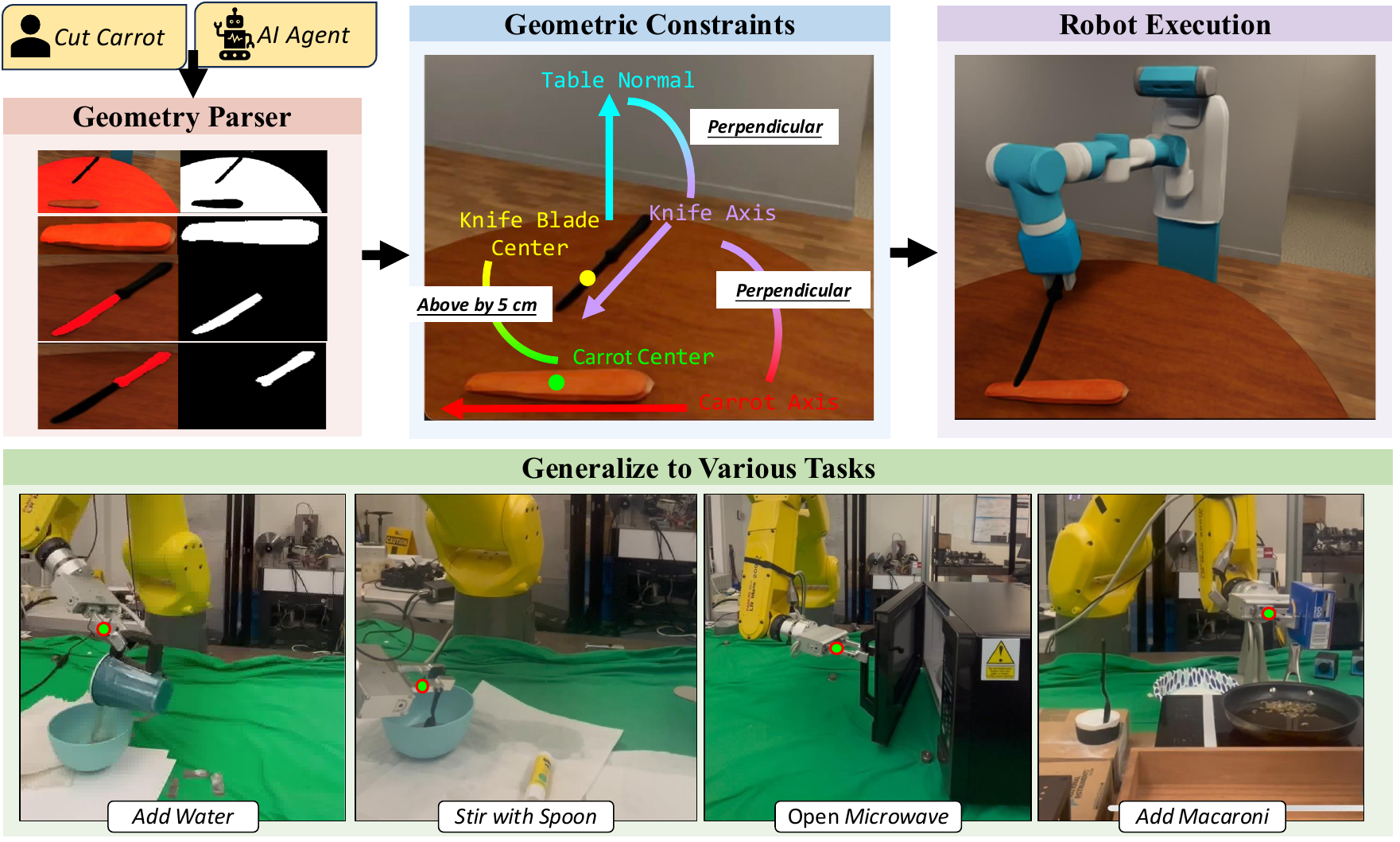}
    \vspace{-6pt}
    \caption{
    We propose to derive geometric constraints to bridge the gap between high-level language descriptions and low-level robot actions. As demonstrated in our results, our \name{} is able
    to execute diverse tasks in general settings. 
    }
    \label{fig:teaser}
    \vspace{-6pt}
\end{figure}

\section{Introduction}
%
%
Recent research~\cite{tang2024embodiment,xu2024flow, ko2023learning, du2024learning, bharadhwaj2024track2act, yuan2024general, baker2022video,chen2021learning, huang2024rekep, liang2023code, huang2023voxposer, duan2024aha} on utilizing vision-language models (VLMs) to develop general robot manipulation policies has drawn much attention, leveraging their vision understanding~\cite{tang2024embodiment,xu2024flow, ko2023learning, du2024learning, bharadhwaj2024track2act, yuan2024general, baker2022video,chen2021learning} and language reasoning~\cite{huang2024rekep, liang2023code, huang2023voxposer, duan2024aha} abilities. 
Such language-based methods offer benefits like providing rich contexts for generalizable and interpretable policies, enabling step-by-step reasoning, and allowing on-the-fly modifications for robotic control. However, language remains conceptual and lacks inherent 3D geometry information, making it challenging to generate precise low-level robot actions.

Vision-language-action models (VLAs)~\cite{kim24openvla,liu2024rdt} have emerged as recent solutions that implicitly bridge perception, reasoning, and execution in an end-to-end manner, enabling robots to plan low-level actions to follow human instructions with contextual awareness.
However, they rely on large-scale and task-specific data for training and often struggle to generalize to novel tasks or objects. Besides, the language-to-action conversion is a ``black box'' without interpretability.
To solve these challenges, a natural solution is to explicitly model the environment and 3D geometry by developing an intermediate representation that can be articulated in a high-level language while also accurately depicting low-level actions, effectively connecting natural language and robotic action.

%
%
%

To this end, we introduce geometric constraints as an interface to connect language instructions with precise robot actions.
With the help of reasoning in vision-language models (VLMs), these constraints can be defined by natural language and then converted into symbolic forms that are able to accurately guide low-level trajectories to complete the task.
%
%
%
See Figure~\ref{fig:teaser} for an example, in the task ``cut the carrot with the knife'', to lift the knife above the carrot, the geometric constraints are: (i) the heading direction of the knife blade must be parallel to the table surface (perpendicular to the normal of the table surface); (ii) perpendicular to the carrot's axis; and (iii) the center of the knife must be positioned approximately 5 centimeters above the center of the carrot. 

In this work, we present \name{}, a framework for building generalist robots by leveraging geometric constraints as an interface to generate precise manipulation trajectories. 
Given a task described in natural language and the current scene observation, \name{} decomposes the task into sub-tasks, ensuring that geometric constraints are satisfied within each sub-task and transitions between sub-tasks are consistent and smooth.
Specifically, \name{} comprises (i) a geometry parser that identifies object parts where geometric properties can be defined, via a proposed \textit{select-process scheme}; (ii) a constraint generator that uses geometric knowledge to produce symbolic constraints and cost functions for each step; and (iii) a cost function-based trajectory solver that minimizes the overall costs, i.e., constraint violation, through optimization.
Notably, code generation is integrated to process selected masks in the geometry parser and represent cost functions in the constraint generator.

With the careful design, \name{} is naturally able to serve as a robot generalist capable of: 
(i) on-the-fly policy adjustments based on human language feedback, (ii) learning from failure cases and adjusting the policy accordingly, (iii) learning from human demonstrations, (iv) performing long-horizon tasks via decomposition, and (v) efficient data collection for imitation learning.
Experiments conducted in both virtual environments, such as MetaWorld~\cite{yu2020meta} and Omnigibson~\cite{li2023behavior}, as well as real-world scenarios, demonstrate \name{}'s wide applicability, training-free, and out-of-distribution (OOD) generalizability across diverse object types, positions, and poses.


%

%
To summarize, our contributions are three-fold:
\begin{itemize}[leftmargin=*]
    \setlength\itemsep{0pt}
    \vspace{-8pt}
    \item We propose a novel generalist robots framework \name{}, using geometric constraints as an interface for robotic manipulation. It is simply driven by high-level language instructions with two key designs: a geometry parser with the select-process scheme and a constraint generator for cost-based planning.
    \item \name{} is capable of reasoning about task constraints with five appealing benefits for robot learning, including learning from human demonstrations, on-the-fly policy adaptation, learning from failure cases, long-horizon planning, and efficient data collection for imitation learning.
    \item Extensive experiments in both simulations and the real world demonstrate the \name{}'s effectiveness and generalizability even to OOD scenarios with no training efforts needed.
\end{itemize}

\section{Related Work}
\noindent \textbf{Robot Manipulation with Large Models.}
There are three branches of work that manage to harness large models for robot manipulation. 
The first branch of work trains or fine-tunes vision-language-action model (VLA) with action-annotated data~\cite{brohan2022rt, brohan2023rt, walke2023bridgedata, ebert2021bridge, huang2023embodied, li2023vision, zhen20243d, driess2023palm, chen2024moto,kim24openvla}, visual affordance data~\cite{li2024manipllm, huang2024manipvqa, yu2024uniaff} or motion tokens~\cite{chen2024moto} 
to achieve end-to-end action predictions given observation.
%
%
The second branch of works~\cite{huang2023voxposer, duan2024manipulate, liu2024moka, huang2024rekep, wang2024dart, liang2023code, ahn2022can, mu2024robocodex, song2023llm, mu2024embodiedgpt, zawalski2024robotic} uses VLM 
as a high-level planner and divide manipulation tasks into sub-goals.
%
The third branch of works ~\cite{huang2023voxposer, huang2024rekep, liang2023code} uses VLM to reason object relations or generate codes to generate low-level trajectory. 
Our work lies in the third branch of work. However, 
different from the closest work ReKep~\cite{huang2024rekep}, we formulate geometric relationships among objects instead of just key point relationships, which depict the manipulation more precisely.
%

\begin{figure*}[t]
    \centering
    \includegraphics[width=0.99\linewidth]{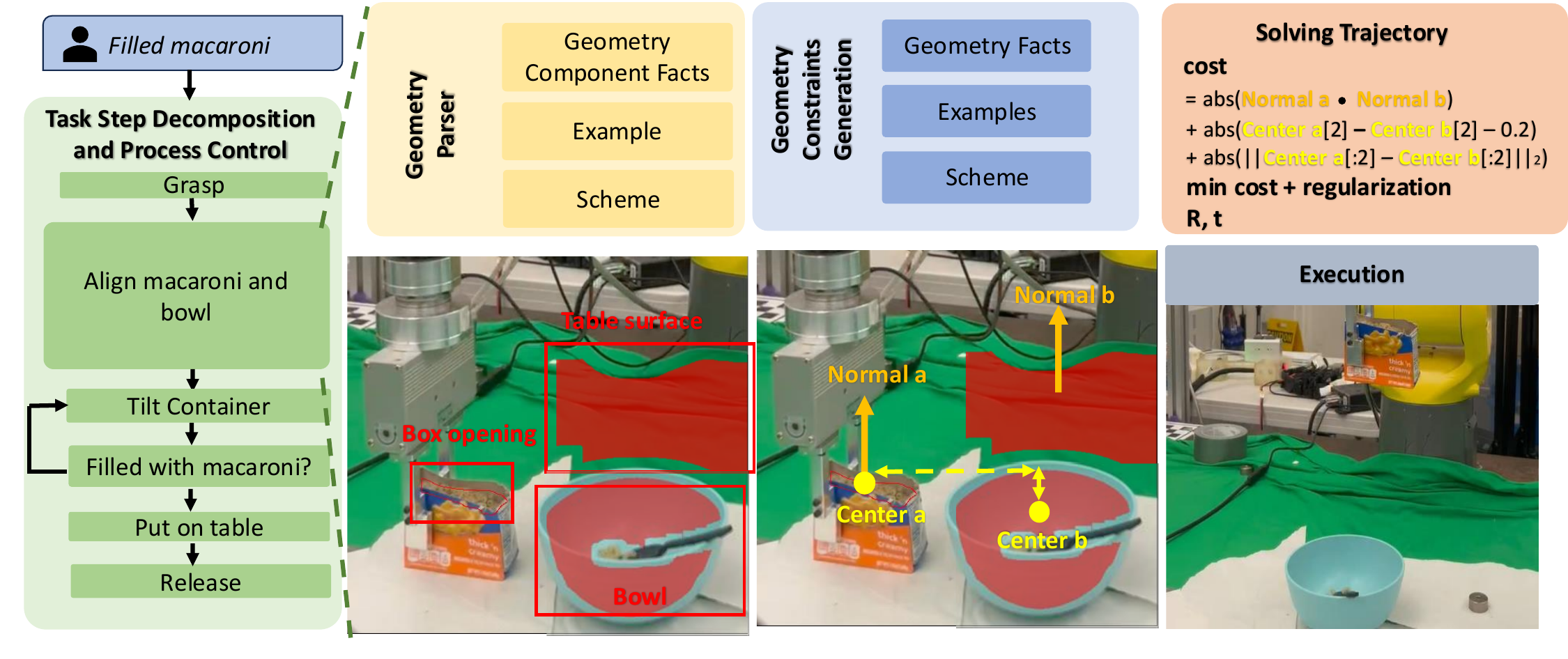}
    \vspace{-12pt}
     \caption{Given the user's task description, our method decomposes the task into multiple sub-tasks and forms the process control. For each stage, we first design a geometry parser to segment and obtain the point cloud for relative geometric components.
     Then, we develop a geometry constraint generation module to generate constraints among the geometric components that are necessary to complete the sub-task.
     Finally, we establish the cost functions to measure the fulfillment of the geometric constraints and solve the robotic trajectories via optimization.}
     \vspace{-12pt}
     \label{fig:framework}
\end{figure*}

\noindent \textbf{Spatial Relation Constraints for Robot Manipulation.}
Spatial relations can be formulated as constraints for robot manipulation. 
Some recent works~\cite{kingston2018sampling, ratliff2009chomp, schulman2014motion, sundaralingam2023curobo, marcucci2024shortest, ratliff2018riemannian} model the manipulation as an optimization problem and solve the constraints globally with various solvers to achieve the desired goal.
%
Some other works
~\cite{toussaint2015logic, toussaint2017multi, toussaint2018differentiable, xue2024d} incorporate multi-stage spatial relationship for planning the whole tasks. 
Recently, Huang et.~\cite{huang2023voxposer, huang2024rekep} use the VLM to reason object spatial relation is automatically given task description. 
Some works~\cite{zhen20243d, zawalski2024robotic} simplicity capture the spatial relationship among objects during model finetuning or inferring, which empowers the model with spatial awareness and improves performance for action prediction.

\noindent \textbf{Open-vocabulary Object Detection and Part Segmentation.}
Vision tasks in an open-vocabulary setting have been challenging because of data limitations. Despite their different model designs and training pipeline, most of the work~\cite{gu2021open, cheng2024yolo, du2022learning, kuo2022f, wu2023aligning, zhong2022regionclip} addresses the open-vocabulary object detection by aligning text embeddings and visual features of local regions.
%
Open-vocabulary part segmentation is even more challenging for the countless part categories. 
The first branch of methods~\cite{wei2024ov, sun2023going, ding2023visual} also makes use of the similarity of semantic and visual features. 
Another branch of methods~\cite{lai2024lisa, zou2023generalized, wang2024llm} finetune the VLM model with part-level segmentation annotation data and harness its reasonability. 
However, these models still cannot achieve satisfactory performance for OOD object part segmentation.

\section{Methods}  

%
Given the current scene observation and human language instructions, we propose the GeoManip framework to utilize the geometric constraints among objects as an interface to generate manipulation trajectories to accomplish the task.
Please refer to Figure~\ref{fig:framework} for an illustration.

The GeoManip framework consists of four steps.
First, we decompose the task into multiple sub-tasks that are completed step-by-step.
For more complex tasks, we create a process control based on these subtasks.
Second, for each sub-task, we present a novel \textit{select-process} solution to identify the relevant geometric components which are fine-grained object-part-level structures that help define the spatial relationships among objects.
%
%
%
Third, we generate the geometric constraints needed to accomplish the task based on the identified geometric components and some fundamental geometric principles we provide.
Finally, we convert the geometric constraints into cost functions that guide the planning of robotic manipulation trajectories through an optimization method.
Note that since the first and the third steps can be learned in an in-context fashion, they can be generated interactively and provide us with 5 important features of a robotic manipulation agent. 

In the following, we introduce the task decomposition and process control in Sec.~\ref{subsec:PC}, present the geometric parser in Sec.~\ref{subsec:CIM}, and detail the constraint generation module in Sec.~\ref{subsec:CGM}.
Finally, we present the cost function and trajectory generation process in Sec.~\ref{subsec:CFG}. We further present the method to achieve 5 generalist features in Sec.~\ref{sec: AI-agent}


\subsection{Task Decomposition and Process Control}
\label{subsec:PC}
An illustration of our process control is shown in Fig.\ref{fig:framework}.
We leverage the VLM's capability for task composition to divide a task into multiple sub-tasks. 
For example, the task ``Filled macaroni" can be decomposed into six sub-tasks.
%
%
For many simple tasks, the sub-tasks are executed sequentially till the end of the task.
%
%
However, complex tasks require loop or branch control.
For example, to pour a certain amount of water from a cup into a container, we should repeatedly tilt the cup and check if the target container has enough water until the desired amount is reached.
This motivates us to add process control during task decomposition. 
To achieve this, we ask the VLM to judge the next sub-task to transit upon finishing the current one.
For example, to ``check if the pan is filled with macaroni'' we allow the VLM to capture previous RGB images to determine if conditions are met.
%

\subsection{Geometry Parser}
\label{subsec:CIM}
Using the language instruction and the current scene observation as input, we identify the fine-grained geometric components of objects. A geometric component is a part of the object on which a geometry can be clearly defined. For example, ``cup opening'' is a geometric component where we can clearly define the plane across it, and ``spoon tip'' is another geometric component where we can clearly define its center point.
These geometric components are necessary for geometric constraint analysis.
%
%
%
%

\begin{figure}
    \centering
    \includegraphics[width=0.99\linewidth]{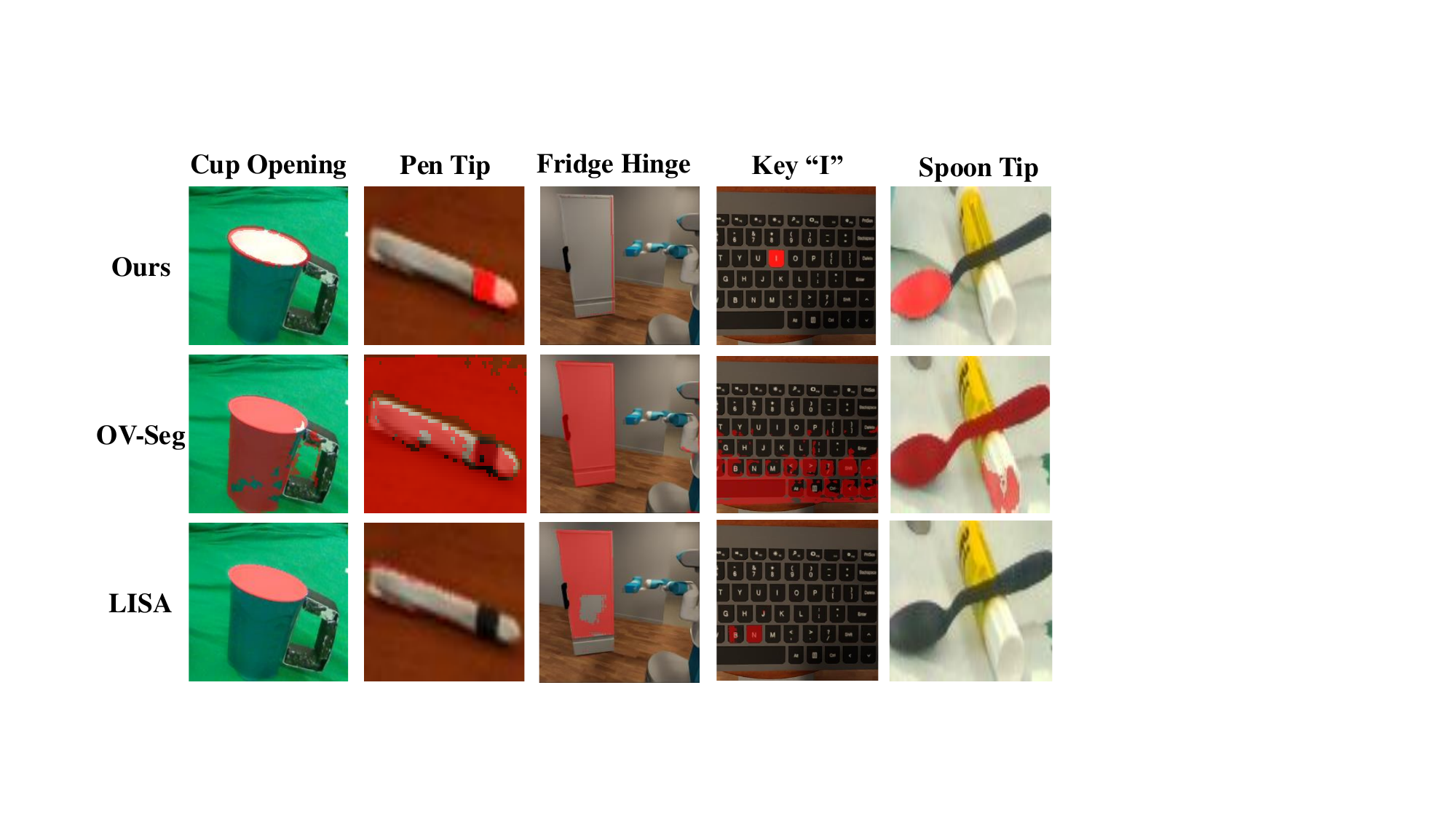}
    \vspace{-6pt}
    \caption{Existing open-vocabulary image segmentation methods (LISA~\cite{lai2024lisa}, OV-seg~\cite{liang2023open}) fail to segment the fine-grained geometric components, while our method segments them correctly.}
    \vspace{-12pt}
    \label{fig:segment_compare}
\end{figure}

However, all existing open-vocabulary image segmentation methods~\cite{lai2024lisa, liu2023grounding, wei2024ov} fail to identify the geometric component.
They may output the entire object or an incomplete object part as illustrated in Fig.~\ref{fig:segment_compare}.
%

Observing that it is relatively easier to perform class-agnostic segmentation and that existing VLMs have an extraordinary ability to understand visual concepts, we combine these two to introduce a select-process scheme to tackle the geometry parsing task.

\begin{figure}
    \centering
    \includegraphics[width=0.99\linewidth]{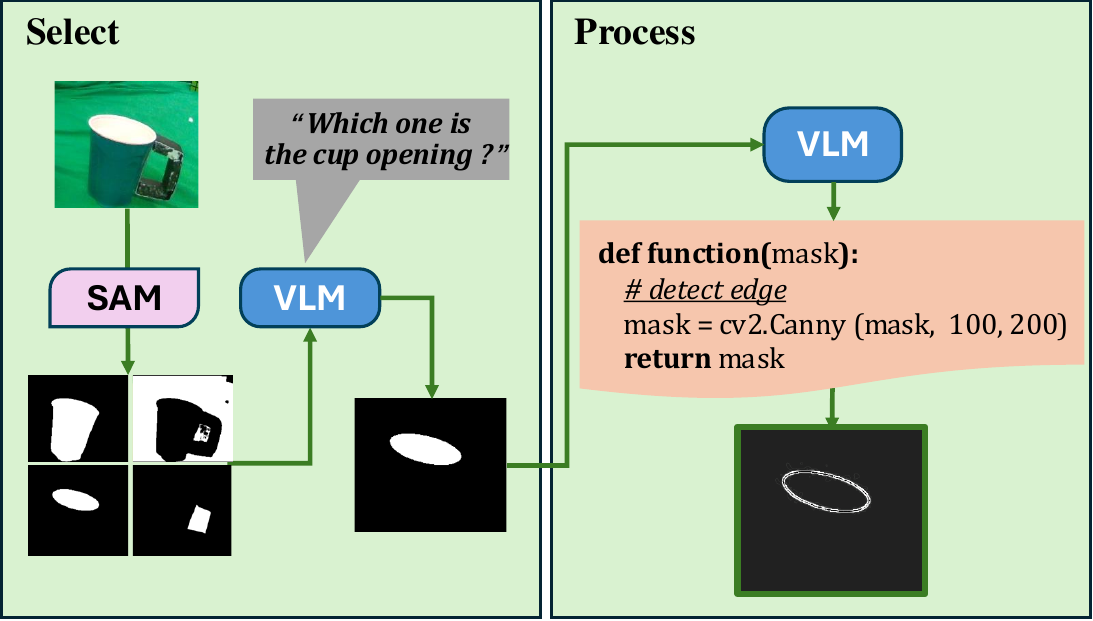}
    \vspace{-6pt}
    \caption{Illustration of our select-process scheme in parsing the geometry.}
    \vspace{-6pt}
    \label{fig:segm-process}
\end{figure}
%

The select-process scheme consists of two steps, as shown in Figure~\ref{fig:segm-process}.
First, we capture an image $I$ of the scene and leverage the Segment-Anything Model (SAM)~\cite{kirillov2023segment} to obtain the class-agnostic masks, i.e., $\{M\}_1^{N} = SAM(I)$.
%
%
%
We further query the VLM to select the most accurate mask. 
Specifically, we provide a language description of the geometric component and pair the image $I$ with each mask to form $\{(I, M)\}_1^{N}$ to aid the selection.
%
%
%
%
However, even after selecting the best-matched mask, it may not accurately represent the geometric component.
Hence, we further leverage the VLM to refine the selected mask to represent the geometric component. 
Let $M^*$ be the selected mask, we use $M^*$, its corresponding image $I^*$, and the geometric part description to query the VLM to implement a code function $g: \mathbb{R}^{H\times W}\rightarrow \mathbb{R}^{H\times W}$ to generate processed mask $M'=g(M^*)$. Please refer to Figure~\ref{fig:segm-process} for an illustration of the processing procedure.

We observe that our method can generate more accurate segmentation results to represent the geometric components as illustrated in Figure~\ref{fig:segment_compare}. 
%
%

\subsection{Constraint Generator}
\label{subsec:CGM}
The constraint generation module infers the geometric constraints among geometric components that are required to complete the current sub-task.
The geometric constraints are defined on the geometric components and describe the spatial relationships among components, e.g., parallel, perpendicular, directly above, to the left by 10 cm, etc.
%
%
%
%
%


%
%
Given the set of geometric components ``\{GeoComp 1, GeoComp 2, $\dots$\}'' involved in the sub-task and a language description of the constraint ``ConsDesc'', a geometric constraint can be formulated as a tuple $($ \{GeoComp 1, GeoComp2, $\dots$\}, ConsDesc$)$.
%
%
For example, for the stage to align a knife with a carrot ready to be cut, 
the geometric constraints are: 
\begin{itemize}[leftmargin=*]
\vspace{-8pt}
    \setlength\itemsep{0pt}
    \item $($\{``the knife blade", ``the carrot"\}, ``the heading of the knife blade is perpendicular to the axis of carrot""$)$
    \item $($\{``the knife blade", ``the table surface"\}, ``the plane of the knife blade is perpendicular to the plane of the table surface""$)$
    \item $($\{``the knife",``the carrot"\}, ``the center of the knife is directly above `the center of the carrot by 10 cm""$)$ 
\end{itemize}

%
%
%
Following~\cite{huang2024rekep}, we further specify if the constraint is a sub-goal constraint (needs to be satisfied only at the end of the trajectory), or a path constraint (satisfied throughout the entire trajectory).

We harness the strong language reasoning ability of VLM to generate geometric constraints automatically. 
%
To achieve this, we need three types of components in the prompt. 
The first is the geometry principles which include some basic geometry facts such as ``To be perpendicular to a plane is to be parallel to its normal''. 
The second is the output rules indicating how the geometric constraint should be formulated.
%
Finally, we include some concrete examples for the VLM to follow. 
Details about the prompts are shown in the Appendix.

\subsection{Cost Functions and Trajectory Generation}
\label{subsec:CFG}
We develop cost functions to quantify the satisfaction of geometric constraints during robotic manipulation, which are used to guide adjustments to the poses of manipulated objects to complete the sub-task.
%
%
Therefore, we propose to use the VLM to generate code that represents the cost functions based on the language format of the geometric constraints, leveraging the established code generation ability of the VLMs.

%
%
More specifically, we ask the VLM to generate a code function $f:\mathcal{P}\rightarrow\mathbb{R^{+}}$ for each cost constraint.
The function $f$ takes the set of geometric component's point clouds $\mathcal{P}=\{\mathbf{p}_1, \mathbf{p}_2, \cdots\}$($\mathbf{p}_i\in \mathbb{R}^{N_i\times 3}$ which is the point cloud of geometric components $i$) as input and outputs a non-negative floating value representing the degree of violation with the geometric constraint (lower is better), and the minimum value of 0 is reached when the geometric constraint is perfectly satisfied.
%

For the VLM to generate the function correctly, we need to provide three components in the prompt as follows:
1. The rules and format for the output. 2. Examples of (geometric constraint, and cost function). 3. General basic geometric facts such as how to orbit or rotate points around an axis. 
The details of the prompt can be viewed in the Appendix.
We obtain a cost function for every geometric constraint, forming a set of path cost functions $\mathcal{F}^p$ for the path constraints and a set of sub-goal path functions $\mathcal{F}^s$ for the sub-goal constraints.

We leverage these cost functions to guide robotic motions by generating the manipulation trajectories to satisfy the geometric components.
First, we identify which geometric component is manipulated by the robotic gripper by finding the ones belonging to the grasping object, we denote the set of point clouds of moving components as $\mathcal{P}^m$.
%
%
%
We further denote the set of stationary geometric components' point clouds as $\mathcal{P}^s$.
Since the gripper is rigidly attached to the moving component $\mathcal{P}^m$, they share the same transformation.
%
Hence, solving the gripper's target 3D rotation matrix $\textbf{R}\in SE(3)$ and transformation vector $\mathbf{t}\in\mathbb{R}^{3}$ for the sub-goal constraints is equivalent to solving the following optimization problem:
\begin{equation}
\small
    \begin{aligned}
        &\min_{\textbf{R}, \textbf{t}} \frac{1}{K^s}\sum_{f\in{ \mathcal{F}^s}} f(\mathcal{P}^s\cup( \textbf{R}\textbf{R}_0^{-1}\bigotimes(\mathcal{P}^m\bigoplus - \textbf{t}_0)\bigoplus\textbf{t})) \\
        &+ \alpha\|\textbf{t} - \textbf{t}_0\|_2 + \beta \|euler(\textbf{R}\textbf{R}_0^{-1})\|_1,
    \end{aligned}
    \label{eq:optimize}
\end{equation}
where $\textbf{R}_0$ and $\textbf{t}_0$ are the gripper previous rotation matrix and translation vector respectively, while $\textbf{R}$ and $\textbf{t}$ denote the optimized rotation matrix and translation vector. $euler(\cdot)$ is the operation to get the Euler angle in three rotation axes from the matrix. $\bigoplus$ and $\bigotimes$ are vector addition and matrix product for each element in the set $\mathcal{P}^m$.
$\alpha$ and $\beta$ are two scalars to regularize in translation and rotation, respectively.
%
After the target rotation $\textbf{R}$ and the translation $\textbf{t}$ of the gripper are obtained, 
we further extract the entire manipulation trajectory.
We first
interpolating between ($\textbf{R}_0$ and translation $\textbf{t}_0$) and ($\textbf{R}$ and translation $\textbf{t}$) and generating several ``control points" between. 
For each ``control points'', we optimize it using $\mathcal{F}^p$ in a similar way as Eq.\ref{eq:optimize}.
%

\subsection{Generalist Embodied Agent}
\label{sec: AI-agent}
Since our method uses geometric constraints as the interface to bridge the high-level planning and the low-level action, it can further be used to develop a generalized embodied agent for robotic manipulation.
See Fig.~\ref{fig:AI-agent} for an illustration of our embodied agent interface, which consists of a user input block, a geometric constraint block, a cost function block, a geometric component visualizer, and a trajectory visualizer.
To accomplish a sub-task, the user uploads an image of the current observation of the scene, together with a language command guiding the agent to generate geometric constraints and the cost function.
Our embodied agent facilitates open-ended conversations with the user while generating geometric constraints, enabling users to provide further instructions.
Video inputs are also allowed so that the agent can learn geometric constraints or observe failures from them.
The agent also visualizes the geometric components and the planned trajectory.
These designs empower \name{} with five features for robotic manipulation:
%
%
\begin{figure*}
    \centering
    \includegraphics[width=0.9\linewidth]{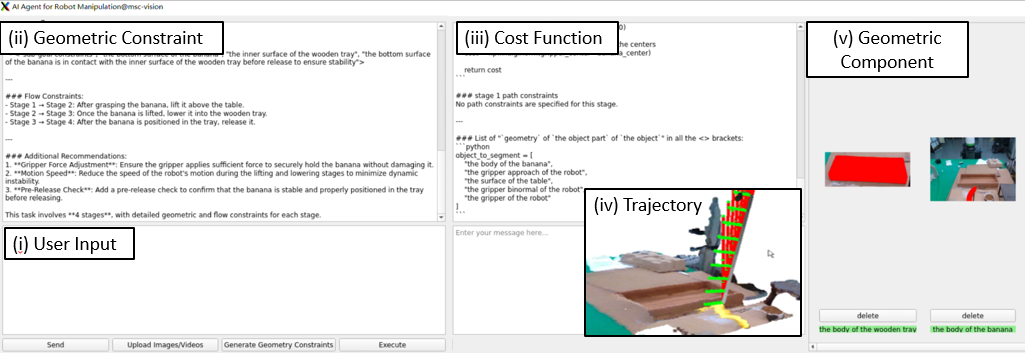}
    \caption{
    Our embodied agent comprises five components: (i) a user input block that accepts the current observation of the scene, the language command from user and uploaded videos of robotic to human manipulation of the sub-task; (ii) a geometric constraint block to display the generated geometric constraints for the sub-task allowing for modifications; (iii) a cost function block to present the developed cost function based on the geometric constraints; (iv) a geometric component visualizer to show the mask of the geometric component involved in the sub-task; (v) a trajectory visualizer that illustrates the planned trajectory in the scene.
    %
    %
    %
    }
    \label{fig:AI-agent}
\end{figure*}

\noindent\textbf{On-the-fly policy adaptation.} 
Since our embodied agent enables the user's open-ended conversation in generating the geometric constraints, the user can augment with further demands to adjust the geometric constraints.
%
For example, the user can specify the distances or rotation angles when manipulating the object in the language command or the user can input high-level commands, such as ``open the door a little bit".
Then, in the geometric constraint block, the embodied agent gives responses to the user's input and also adjusts the geometric constraint.
the cost functions and the manipulation trajectory are adjusted at the same time.

\noindent\textbf{Learn from failure cases.}
Our method can learn from previous failure executions and improve the current manipulation policy.
To achieve this, the user uploads recordings of the robot's failed execution to the embodied agent, accompanied by language commands like "Why did the robot fail to execute?" and "How can we adjust the geometric constraints to improve performance?".
The embodied agent then re-generates the geometric constraints.
%
%

\noindent\textbf{Long-horizon task planning.}
Our embodied agent can also deal with long-horizon tasks.
%
%
Specifically, we first ask the agent to divide the long-horizon tasks into multiple single tasks The results are displayed in the geometric constraint block, together with the geometric constraints for each short-horizon task.
%
The agent then deals with each single task one by one.

\noindent\textbf{Learn from human demonstrations.}
Our embodied agent can receive manipulation videos from humans or the robot in the user input block by asking it to generate geometric constraints according to the videos.
%
These learned extracted constraints can be generalized to similar objects.

\noindent\textbf{Data collection for supervised model training.}
Our method not only produces a directly executable policy for completing the task but also functions as a data collection strategy, contributing to the training of other methods.
Specifically, we collect the trajectory generated by our embodied agent together with the visual input of the scene and use them as the ground truth to supervise the vision-language-action model.
Also, the generated cost function can be viewed as a reward function as it inherently reflects how close the current robotic state is to the target state for the manipulation task.
Therefore, the cost function can also be used to train a reward model.

\section{Experiments}
\label{sec:exp}
\subsection{Implementation Details}

\noindent \textbf{Technical Details for the VLM design.}
We use GPT-4o~\cite{openai2024chatgpt} as the VLM for our implementation. We use the SLSQP algorithm~\cite{kraft1988software} for optimization solving. For optimization, we set $\alpha=0.02$, and $\beta=0.075$ for regularization. We use Grounding-DINO~\cite{liu2023grounding} to locate and crop the target object first before processing it with the geometry parser to prevent overwhelming mask candidates generated by SAM~\cite{kirillov2023segment}.

\noindent \textbf{Virtual Benchmarks.} We perform experiments on two virtual environments: Meta-World~\cite{yu2020meta} and OmniGibson~\cite{li2023behavior}, including 10 diverse tasks.
The Meta-World environment is a simulated benchmark featuring numerous predefined tasks for reinforcement learning.
We compare the performance of our method on 6 tasks, following the common settings~\cite{tang2024embodiment, ko2023learning}.
The OmniGibson environment is another virtual environment.
It features realistic physics simulation and rendering and enables user-defined tasks.
%
%
%
We further develop 4 tasks on OmniGibson to test our method.

\noindent \textbf{Real-world Environment.}
In addition to the experiments on the simulators, we design 4 more tasks to demonstrate the effectiveness of our method in real-world robotic settings.
A table is set up with objects placed on top of it.
A RealSense D435i camera is set up for visual sensing and a FANUC LR mate 200id robot arm is equipped to perform the task.

\noindent \textbf{Evaluation Metrics.} 
For all benchmarks, we assess the performance of all methods based on each task's success rate (number of success trials/number of total trials).
%

\subsection{Results on Virtual Benchmarks}
\noindent \textbf{Meta-World Environemnt.}
We first evaluate the performance of the methods on 6 tasks in the Meta-World benchmark~\cite{yu2020meta} and the results are shown in Table~\ref{table:mw_main}.
We follow the common settings~\cite{tang2024embodiment, ko2023learning} and only consider gripper translation in this environment.
%
%
%
%
%
We use the ground-truth mask for objects that are too small to be identified by the VLM. We test each task for 3 camera positions and report the best result across different views. 
%
%
For each task and each camera view, we evaluate our method 5 times, and at the start of each trial, we randomize the initial poses and positions of the object involved in the task.
We report the success rates and provide an overall success rate across all tasks.
The detailed settings can be viewed in the Appendix.

\begin{table*}
\centering
\small
\caption[]{\textbf{Results on the Meta-World Dataset.}}
\vskip 0.1in
\setlength{\tabcolsep}{9.1pt}

\resizebox{0.99\linewidth}{!}{\begin{tabular}{lcccccccc}\toprule
& basketball & shelf-place & btn-press & btn-press-top & handle-press & assembly & \textbf{overall} \\
\midrule
BC-Scratch & 21.3\% & 36.0\% & 0.0\% & 0.0\% & 34.7\% & 0.0\% & 15.3\%\\
BC-R3M & 0.0\% & 0.0\% & 36.0\% & 4.0\% & 18.7\% & 0.0\% &  9.8\% \\
UniPi & 0.0\% & 0.0\% & 6.7\% & 0.0\% & 4.0\% & 0.0\% & 1.8\% \\
Diffusion Policy & 8.0\% & 0.0\% & 40.0\% & 18.7\% & 21.3\% &  1.3\% & 14.8\% \\
AVDC & 37.3\% & 18.7\% & 60.0\% & 24.0\% & 81.3\% & 6.7\% & 38.0\% \\
SceneFlow & \textbf{96.0}\% & 29.3\% & 50.7\% & \textbf{96.0}\% & 40.0\% & \textbf{46.7}\% & 59.8\% \\
\midrule
\textbf{Ours} & 73.3\% & \textbf{60.0}\% & \textbf{80.0}\% & 73.3\% & \textbf{100.0}\% & 40.0\% & \textbf{71.1}\% \\
\bottomrule
\end{tabular}}
\label{table:mw_main}
\end{table*}

We compare our method with six state-of-the-art methods, i.e., BC-Scratch~\cite{nair2022r3m}, BC-R3M~\cite{nair2022r3m}, UniPi~\cite{du2024learning}, Diffusion Policy~\cite{chi2023diffusion}, AVDC~\cite{ko2023learning}, and SceneFlow~\cite{tang2024embodiment}.
Note that all of the existing methods require an additional training stage to learn the robotic action, while our method is training-free.
%
%
%
The results of our method and the compared methods are presented in Table~\ref{table:mw_main}.
From the table, we can see that our method greatly outperforms BC-Scratch, BC-R3M, UniPi and AVDC and SceneFlow by over 11\% in terms of average accuracy, while our method does not need any training stages.
The results demonstrate the effectiveness of our method.
%
%
%

\noindent \textbf{OmniGibson Environment.}
What's more, we evaluate our method on 4 tasks, i.e., opening the fridge, typing the requested letter on the keyboard, putting a pen into a pen-holder, and cutting a carrot with a knife.
These tasks require both the translation and rotation of the gripper built on the OmniGibson to further demonstrate the effectiveness of our method.
%


%
%
%

We conduct 5 trials for each task and report the success rates to evaluate the methods.
We also compare our method with a very recent work evaluated on the OmniGibson environment, the ReKep~\cite{huang2024rekep}. 
The ReKep proposes to plan robotic manipulation based on the spatial relations among the key points on objects.
The experimental results of our method and the compared method on the OmniGibson is shown in Table~\ref{table:mw_main2}.
From the table, we can see that our method consistently outperforms ReKep in each task.
It outperforms ReKep by at least 20\% in each task and achieves a 40\% higher overall success rate.
%
%
This is because our method better models the relations between objects via geometric constraints, which is more detailed and precise compared with the object's key points.


\begin{table}[t]
\centering
\small
\caption[]{\textbf{Results on the Omnigibson Environment.}}
\vskip 0.1in
\setlength{\tabcolsep}{3.1pt}
\resizebox{\linewidth}{!}{
\begin{tabular}{lccccc}\toprule
& open-fridge & typing & put-pen-into-holder & cut-carrot & \textbf{overall}\\
\midrule
ReKep & 0.0\% & 0.0\% & 60.0\% &  20.0\% & 20.0\%\\ 
\textbf{Ours} & \textbf{80.0}\%  & \textbf{40.0}\% & \textbf{80.0}\% & \textbf{40.0}\% & \textbf{60.0}\%\\
\bottomrule
\end{tabular}}
\label{table:mw_main2}
\vspace{-12pt}
\end{table}




\subsection{Experiments on Real Environment}
Furthermore, we test our method in real-world settings.
We design 4 tasks in the real-world environment, i.e., picking and placing an object onto another object, pouring something from one container to another, opening/closing an object with rotation / prismatic movement, stirring something in a container.
%
%
For each task, we randomize the object type and the initial poses of objects. 
%
%
More details can be found in the Appendix.

\begin{figure*}[t]
    \centering
    \includegraphics[width=\linewidth]{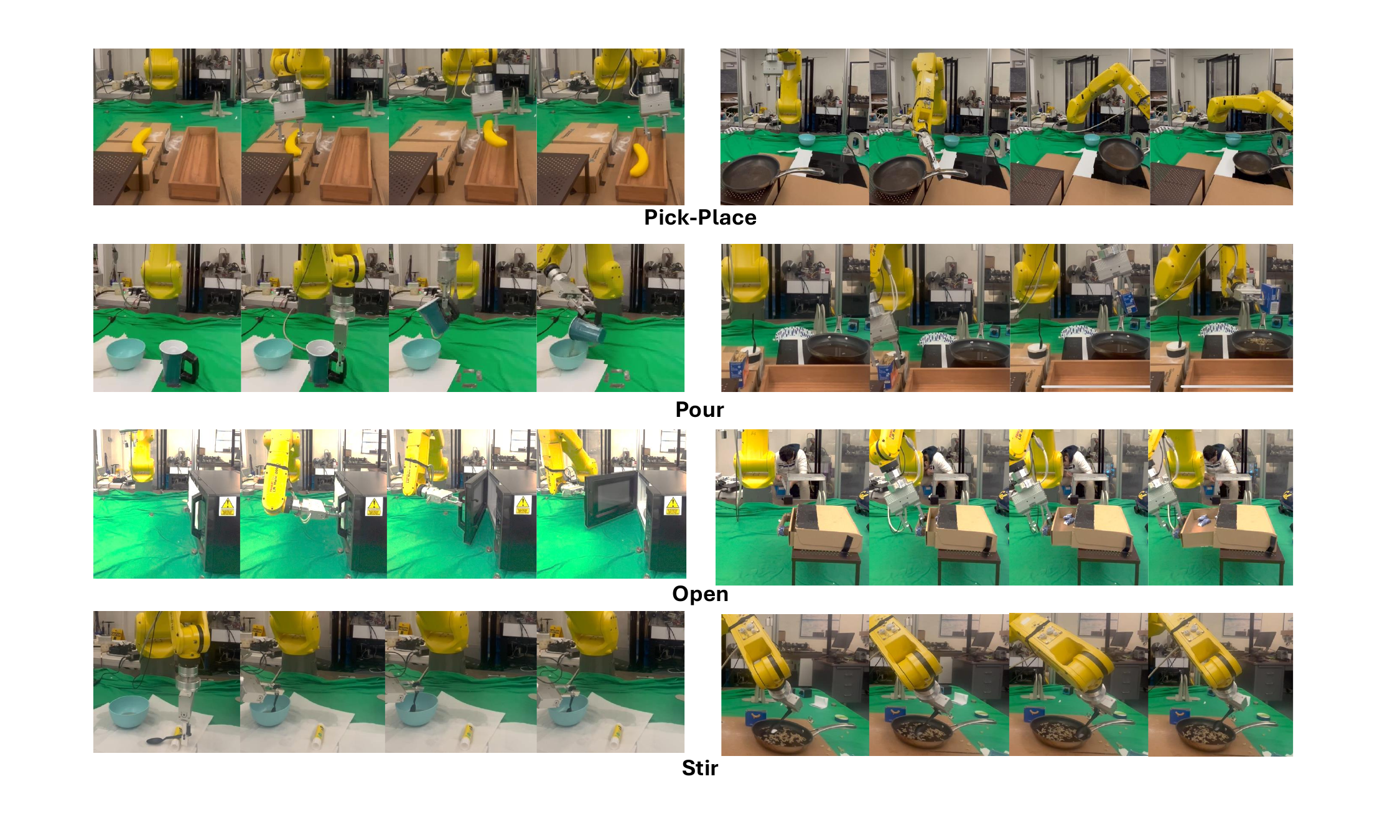}
    \caption{Demonstration sequences and testing scenarios for all four real-world experiment tasks. 
    }
    \label{fig:real-world}
\end{figure*}

We test 10 times for each task and report the success rate. 
%
%
We also established a baseline method,  OpenVLA~\cite{kim24openvla} for comparison. 
It is a training-required behavior-cloning method that leverages a transformer to train the robotic manipulation policy based on visual observation.
%
We collect 30 training trials with manual manipulation in each task for finetuning the OpenVLA. 
The statistical results are shown in Tab.~\ref{table:real-world}.
The results show that even though our method requires no training, it significantly outperforms OpenVLA.
For each task, our method achieves at least a 40\% higher success rate than OpenVLA, leading to an overall success rate that is 50\% greater.
The results also show that by using the geometric constraint as the abstract interface, our method achieves high performance in real-world robotic manipulation and is robust to object poses, locations, types, and tasks.
We visualize some of our demonstration sequences in Fig.~\ref{fig:real-world}. 
From the visualization results, we can see that since our method can understand the geometric constraint, it can successfully manipulate various tasks involving different object types.


\begin{table}[t]
\centering
\small
\caption[]{\textbf{Results on the Real-World Environment.}}
\vskip 0.1in
\setlength{\tabcolsep}{9.1pt}
\resizebox{\linewidth}{!}{
\begin{tabular}{lccccc}\toprule
& pick-place & pour & open & stir & \textbf{overall}\\
\midrule
OpenVLA & 30.0\% & 20.0\% & 10.0\% &  0.0\% & 15.0\%\\ 
\textbf{Ours} & \textbf{90.0}\%  & \textbf{70.0}\% &  \textbf{60.0}\% & \textbf{40.0}\% & \textbf{65.0}\%\\
\bottomrule
\end{tabular}}
\label{table:real-world}
\vspace{-12pt}
\end{table}

\section{Generalist Embodied Agent for Robotic Manipulation}
After demonstrating the effectiveness of our method in Section~\ref{sec:exp}, we further show the five attractive features of our embodied agent. 
Please refer to  Section~\ref{sec: AI-agent} for more details about our embodied agent.

\subsection{On-the-fly Policy Adaptation}
\begin{figure}
    \centering
    \includegraphics[width=\linewidth]{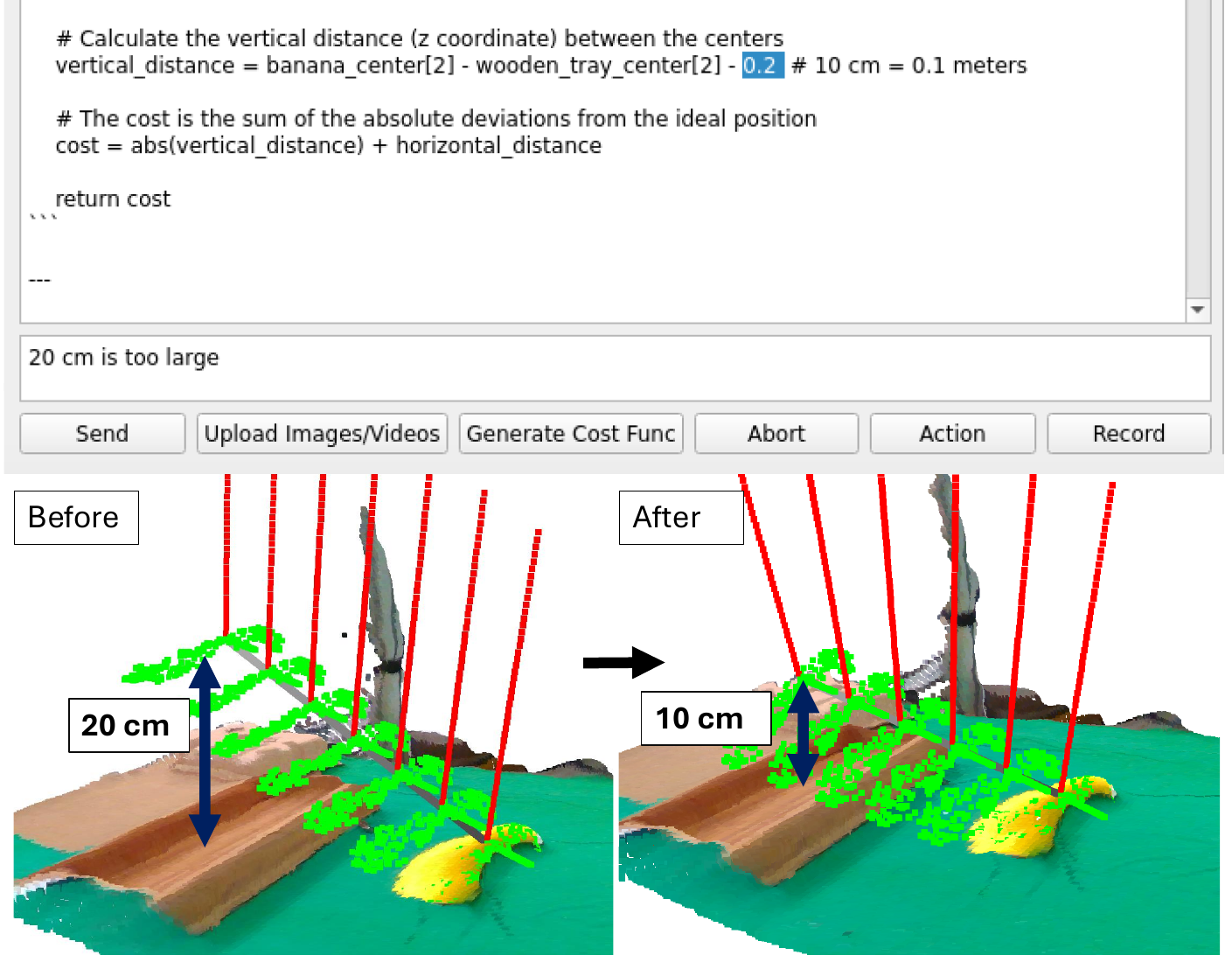}
    
    \caption{Example of the embodied agent adjusting geometric constraint based on human feedback. The red lines are the approaching direction of the gripper, and the green line is the binormal direction of the gripper. The original geometric constraint plans to lift the banana 20 cm above the wooden box as illustrated on the left. By indicating "the height is too large" in the conversation (top of the image), the embodied agent reduces the height and plans to lift the banana from 20 cm to 10 cm in height (right of the image).}
    \label{fig:policy-adaptation}
\end{figure}

%
%
%
By specifying the height above the wooden tray, the robot lifts the banana above the wooden tray with a height modified from 20 cm to 10 cm as illustrated in Fig.~\ref{fig:policy-adaptation}.

\subsection{Learn from Failure Cases}
\begin{figure}
    \centering
    \includegraphics[width=\linewidth]{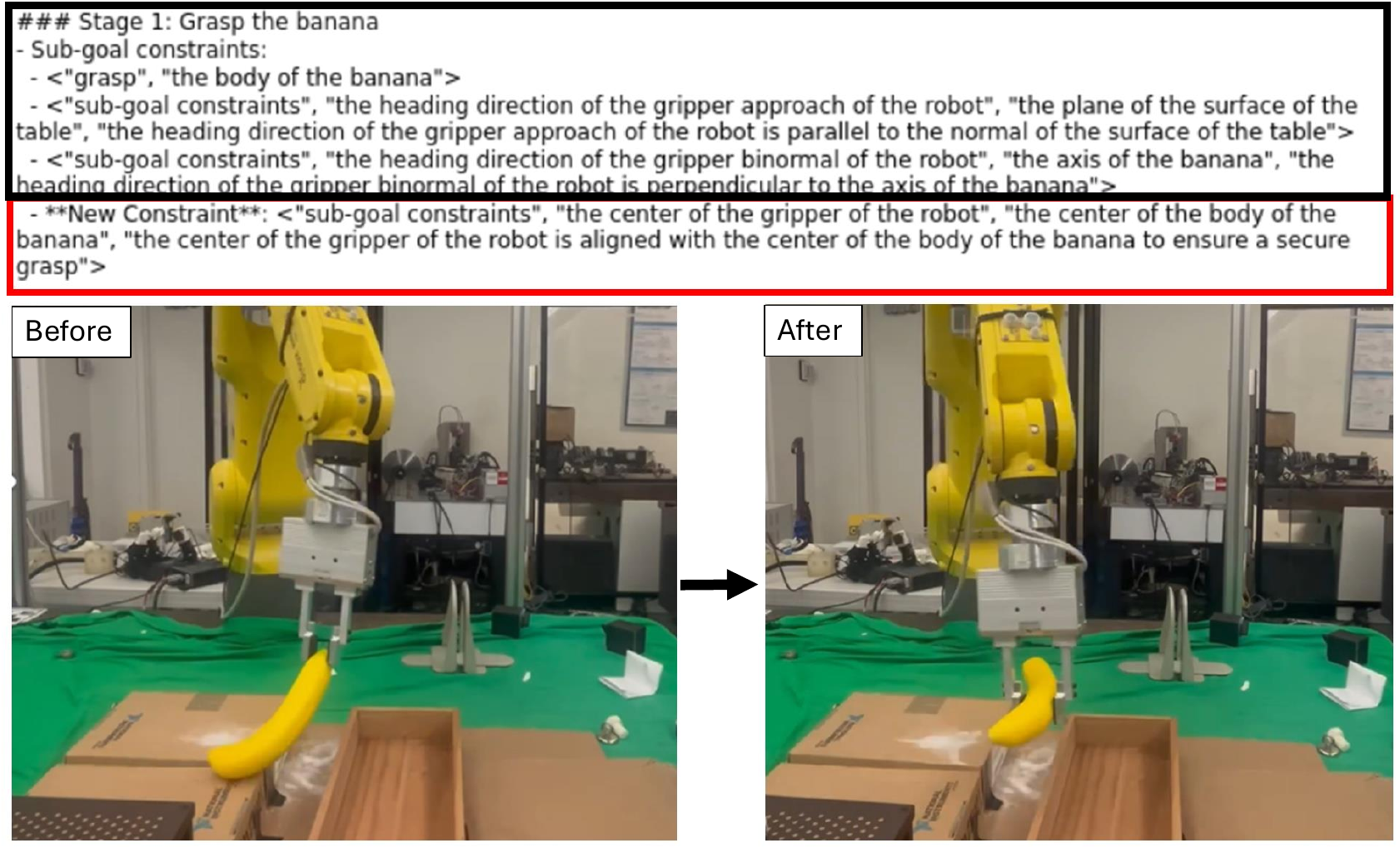}
    \vskip -0.1in
    \caption{Example of the embodied agent learning from the failure case. The original geometric constraints are the first 3 constraints. Under these constraints, as illustrated in the left image, the robot may grasp the banana in an unsafe grasping position. After prompting the embodied agent with the failure video, it can identify the unsafe grasping position and the new constraint is added as highlighted in the red box. With the new constraint, the grasping pose is closer to the banana's center and the task is executed successfully as illustrated in the right image.}
    \label{fig:learn_from_failure}
\end{figure}

%
%
We showcase the ability of our agent to learn from failure cases in Fig.~\ref{fig:learn_from_failure}.
The robot fails to place the banana into the wooden tray because of the unsafe grasp position. 
By asking the embodied agent "The robot fails and the banana slips", it refines the geometric constraint and adds an extra sub-goal constraint during the grasp stage to enforce the grasp position to be close to the center of the banana. 
After re-solving for the trajectory, the banana is safely grasped and transferred to the wooden tray successfully. 

\begin{figure*}
    \centering
    \includegraphics[width=\linewidth]{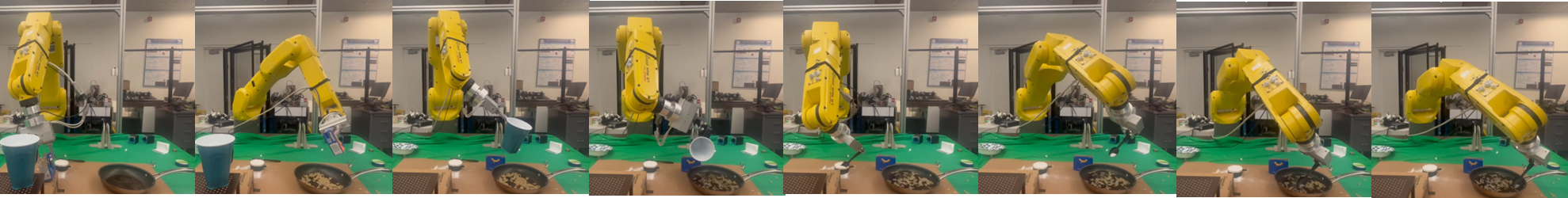}
    \vskip -0.1in
    \caption{Example the embodied agent handling the long-sequence task: ``Add the pan with macaroni and water. Add salt with the spoon and stir the pan.".}
    \label{fig:fig_long_sequence}
\end{figure*}

\subsection{Long-horizon Tasks}
%
We successfully achieved a long-sequence demo for the task ``Add the pan with macaroni and water. Add salt with the spoon and stir the pan." as illustrated in Fig.~\ref{fig:fig_long_sequence}. 
The complete video of the demonstration can be referred to in the Appendix.

\subsection{Learn from Human Demonstrations}
\begin{figure}[t]
    \centering
    \includegraphics[width=\linewidth]{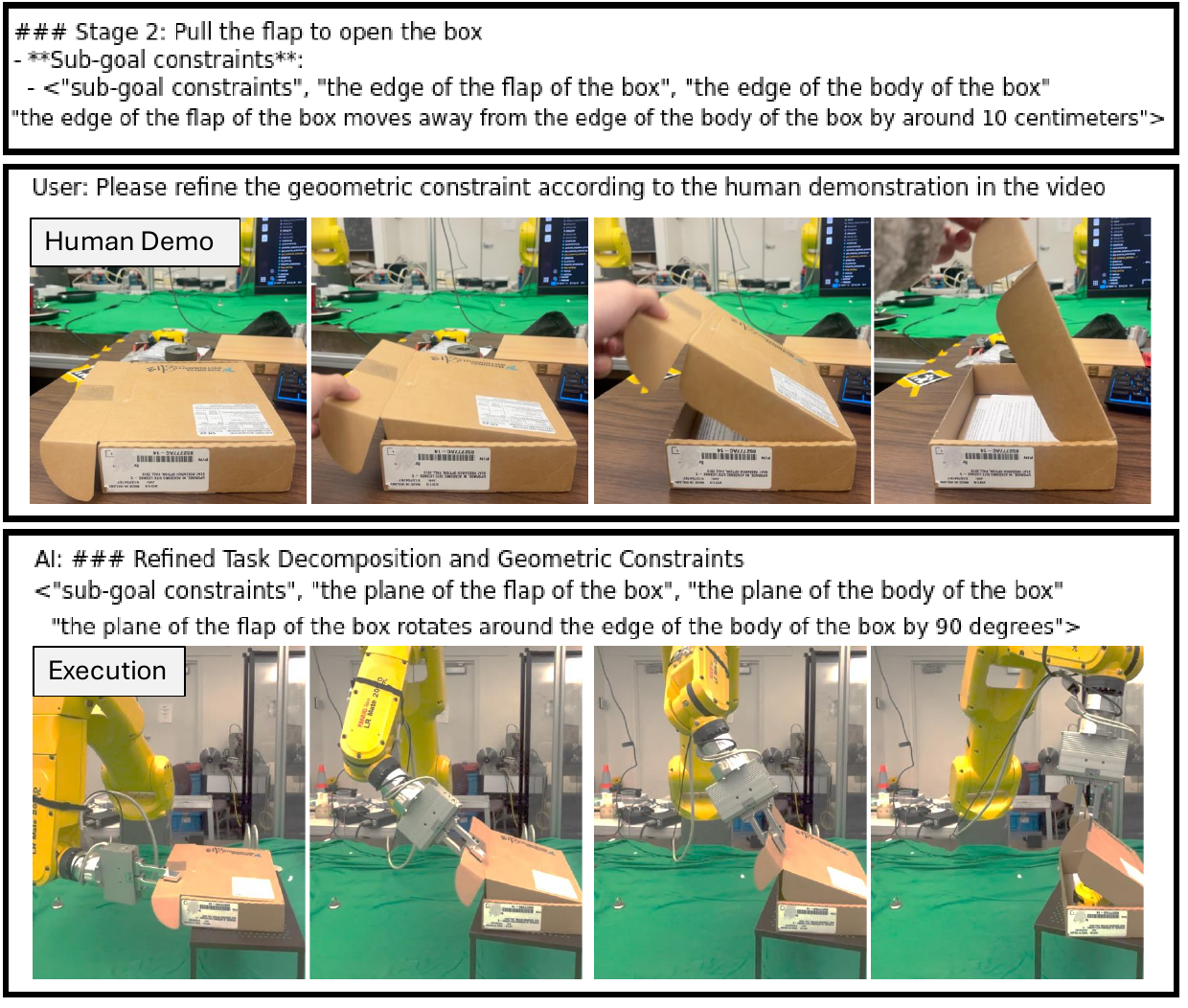}
        \vskip -0.1in
    \caption{Example of the embodied agent learning from human demonstration for the task "open the box". The first line is the original geometric constraint. The second line is the user requirement for learning from the video as well as the human demonstration video. The line below is the refined geometric constraint after learning from a human demonstration, and the image sequence illustrates the execution process. 
    }
    \label{fig:learn_from_human_demo}
\end{figure}
We demonstrate the effectiveness of our embodied agent to learn from human demonstrations.
As demonstrated in Fig.~\ref{fig:learn_from_human_demo}, in the task of ``open the box", the original sub-goal constraint treats the box as the drawer and the policy tries to move the flap of the box away from the box center. 
After including a human demonstration, it correctly captures the box as open by lifting the lid.
Therefore, it refines the sub-goal constraint to lift and rotate the flap around the box edge, resulting in successful box opening.

\subsection{Data Collection for Training Models}
%
In the following, we showcase how our method generates data for training robotic policies.
Specifically, we show the performance of our training data collection scheme for the Vision-Language-Action (VLA) models and the reward models.
%
%
\subsubsection{Data Collection for VLA Model.}
Since the geometric constraints remain consistent for the same manipulation task, even when the positions and orientations of the objects are changed, we only need to generate the geometric constraints once.
 %
 Also, we only need to perform segmentation once at the first object configuration, and then
 use point-tracking models such as CoTracker~\cite{karaev2023cotracker} to track the geometric component for other initial poses and positions.
 %
 Therefore, we can efficiently generate the ground-truth trajectory for varied initial poses and positions quickly, and use it to train the VLA model.
 For the experiment, we apply this strategy to collect data for fine-tuning the OpenVLA model~\cite{kim24openvla} under two tasks: 1. Pick-stick: the robot needs to pick up the wooden stick and place it in the pan. The pan is large enough so that the task can be successfully achieved as long as the stick is above the pan. 2: Pick-banana: the robot needs to pick the banana and place it into a slim wooden box, which can only succeed if the banana axis is aligned with the long side of the wooden box. 
 We collected 30 training data using our strategy, and we compared the performance of OpenVLA fine-tuning on our data and fine-tuning on manually collected ones.
 %
 Note that our primary focus is on learning the manipulation trajectories, so we consistently use the ground-truth grasp pose in both settings to minimize interference from object grasping.
 We can see that the trajectories efficiently collected by our method (Ours) match a similar quality with the ones manually collected (Manual).
 %
 %
 
 \begin{table}[t]
     \centering
     \caption{Performance comparisons between Open-VLA trained with manually collected and data collected with our methods.}
    \label{tab:vla}
     \begin{tabular}{lcccc}
         \toprule
        & Place-stick & Place-banana \\
        \midrule
        Manual & 3/5 & 5/5 \\ 
        \textbf{Ours} & 3/5  & 4/5  \\
        \bottomrule
        \end{tabular}
        \vspace{-22pt}
 \end{table}

\subsubsection{Data Collection for Reward Model.}
\begin{figure}
    \centering
    \includegraphics[width=\linewidth]{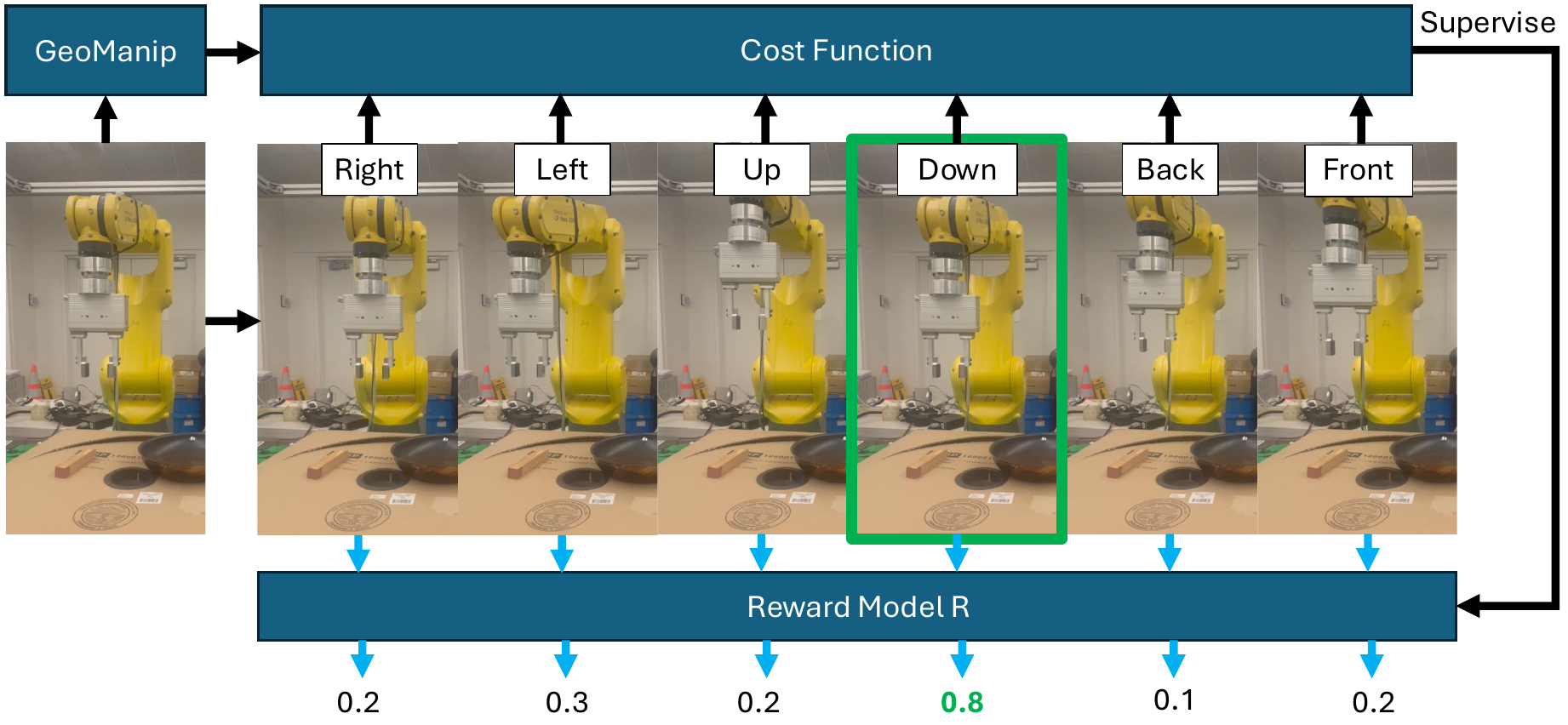}
        \vskip -0.1in
    \caption{Example of using \name{} to train the reward model $R$. The blue arrow is the data flow during inference. We select and execute the action that maximizes the reward (in this case, $a=$``Down'').}
    \label{fig:reward_model}
\end{figure}

Furthermore, the generated cost function can indicate how close the current robotic state is to the target state for the manipulation task, the cost function itself can be viewed as a reward function. 
%
%
%
As a result, we can readily derive the reward value from the cost function to train a reward model that takes the current RGB observation $o$ as input.
%
The model uses a simple ViT model as an encoder to encode the RGB image and uses a Multi-Layer Perceptron (MLP) to generate the reward score. 
During inference, we define a set of candidate actions: \{Left, Right, Front, Back, Up, Down\}, and each action moves the gripper positions along the corresponding direction by a small step. 
By denoting $o'=step(o, a)$ as the RGB observation after applying action $a$ on the previous scene with RGB observation $o$, we select the action $a$ according to $a=\underset{a}{\mathrm{argmax}} R(step(o, a))$. 
We designed a naive example in which the robot needs to pick up the wooden stick. 
%
This concept and process is illustrated in Fig.~\ref{fig:reward_model}.

\section{Limitations}
Our work limitations come with two aspects: geometry parsing and geometric constraint generation. For the former, our method fails when: 1. The geometric component is not clearly shown in the camera. 2. The point cloud in the geometric component is not of good quality. 3. The geometric component is difficult to describe. For the latter, our method fails when: 1. The LVM model misses the necessary geometric constraints to complete the task successfully. 2. The action cannot be clearly described in language. 
We will address these limitations in our future work.

\nocite{langley00}

\bibliography{main}
\bibliographystyle{icml2025}

\clearpage
\appendix
\onecolumn
\section{Prompts}
For each of the Task Decomposition and Flow Control Module, geometry parser, Geometric Constraint Generation Module and Cost Function Generation Module, we design a scheme prompt and an example / knowledge prompt. The scheme prompt provides rules that the VLM model should follow to generate valid output. The example / knowledge prompts provide necessary example to follow or knowledge for in-context learning. For convenience in our implementation, we combine the prompts for Task Decompositions and Flow Control, and Geometric Constraint Generation modules into one, resulting in 6 prompts:
\begin{itemize}
    \item Prompts of schemes for Task Decompositions,  Flow Control, and Geometric Constraint Generation
    \item Prompts of Examples for Task Decompositions, Flow Control, and Geometric Constraint Generation
    \item  Prompts of schemes for Cost Function Generation
    \item Prompts of Geometric Knowledge for Cost
Function Generation
    \item Prompts of schemes for Segmentation Mask Selection and Processing
    \item Prompts of Segmentation Knowledge for Segmentation Mask Selection and Processing
\end{itemize}
\subsection{Prompts of Schemes for Task Decompositions,  Flow Control, and Geometric Constraint Generation}
\begin{spverbatim}
## Query
Query Task: "{Task Description}"

## Instructions
Suppose you are controlling a robot to perform manipulation tasks. The manipulation task is given as an image of the environment. For each given task, please perform the following steps:
1. Task decomposition and flow control: 
Determine how many stages are involved in the task. 
Grasping or releasing must be an independent stage. 
Flow control controls the transition between stages
Some examples:
  - "pouring tea from teapot":
    - 6 stages: "grasp teapot", "align teapot with cup opening", "tilt teapot", "flow control: repeat 'tilting teapot' until the cup if filled", "place the teapot on the table", "release"
  - "put red block on top of blue block":
    - 3 stages: "grasp red block", "drop the red block on top of blue block"
  - "reorient bouquet and drop it upright into vase":
    - 3 stages: "grasp bouquet", "reorient bouquet", and "keep upright and drop into vase"
2. Geometric constraint and flow constraint generation: For each stage except for the grasping and release stage, please write geometric constraints and flow constraint in lines. Each line represent a constraint should be satisfied. 
- Geometric constraint is a tuple of multiple element: <"constraints type", "`geometry 1` of `the object part` of `the object`, "`geometry 2` of `the object part` of `the object`, ...(if any),  "constraints">, each element is explained in the follows:
  - "geometry":  Basic geometric primitive like the left edge, the center point, the plane, the normal, the right area, heading direction, and etc..
  - "the object part": the key object part on an object, like the tip, the opening, the handle, the hinge, the slider, the gripper, etc.
  - "the object": the complete object, like the black cup, the second door, the teapot, the robot, etc.
  - "constraint":
    - 1. basic geometric relationship including parallel, perpendicular, vertical, intersect, and etc.. 
    - 2. positional constraint like above, below, to the left / right, and etc.. 
    - 3. Distance range like "by 10 centimeters", "around 10 centimeters", "more than 25 centimeters", "10 centimeters to 20 centimeters", "45 degress", etc..
    - 4. Transformation like "rotate", "shift", etc.
  - Specify the <`geometry` of `the object part` of `the object`> in the "constraint"
  - "constraints type": 
    1. "sub-goal constraints": constraints among `geometry 1`, `geometry 2`, ... that must be satisfied **at the end of the stage**. In other word, it specifies the constraints of the destination position.
    2. "path constraints": constraints among `geometry 1`, `geometry 2`, ... that must remain satisfied **within the stage**. In other word, it specifies the constaints on the way to the destination position.
- Flow constraint is a tuple of multiple element <"flow constraint", "condition"> (goto stage ? if satisfied; goto stage ? if not satisfied)
- For each stage, there can be ONLY one flow constraint. If there are multiple flow constraint, use standalone stage to place these flow constraints
- Do not ignore "of". There must of at least two "of": "`geometry` of `the object part` of `the object`". If you what to specify `geometry` of the whole object, use `geometry` of the body of `the object`
- For the grasping stage, sub-goal constraint 1 should be  <"grasp", "the area of `the object part` of `the object`">
- For grasping stage, you can also specify the sub-goal constraints of the heading direction of the gripper approach of the robot or the heading direction of the gripper binormal of the robot:
  - approach: the direction of the robot gripper pointing at, usually perpendicular to some surface. You can get the gripper approach by calling get_point_cloud("the gripper approach of the robot", -1). To find its heading direction, find its eigenvector with max eigenvalue.
  - binormal: the direction of gripper opening / closing, usually perpendicular to some axis / heading direction or parallel to some normal. You can get the gripper binormal by calling get_point_cloud("the gripper binormal of the robot", -1). To find its heading direction, find its eigenvector with max eigenvalue.
- To close the gripper only without grasping anything, output <"grasp", "">
- If you want to use the gripper, only specify its center position, the heading direction(approach), or the binormal. 
- For the releasing stage, sub-goal constraint should be <"release">
- Avoid using the part that is invisible in the image like "bottom", "back part" and etc.
- Please give as detailed constraint as possible.
- To move something, you must grasp it first.
- Each stage can only do a single action one time.
- Don't omit stages for the repeating stages, expand and list them one by one.
- Please answer according to the image we provided, which is the previous scene.

\end{spverbatim}
\subsection{Prompts of Examples for Task Decompositions, Flow Control, and Geometric Constraint Generation}
\begin{spverbatim}
Examples for geometric constraint generation and flow constraint generation for each stage under the task:
  - "pouring liquid from teapot until the cup is filled": 
    - "grasp teapot" stage: (stage 1)
      - <"grasp", "the handle of the teapot">
      - <"sub-goal constraints", "the heading direction of the gripper approach of the robot", "the plane of the surface of the table", "the heading direction of the gripper approach of the robot is parallel to the plane of the surface of the table">
      - <"sub-goal constraints", "the heading direction of the gripper binormal of the robot", "the heading direction of the handle of the teapot", "the heading direction of the gripper binormal of the robot is perpendicular to the heading direction of the handle of the teapot">
    - "align teapot with cup opening" stage: (stage 2)
      - <"sub-goal constraints", "the center of the teapot spout of the teapot", "the center of the cup opening of the cup", "the center of the teapot spout of the teapot is directly above the center of the cup opening of the cup around 20 centimeters">
    - "tilt teapot until the cup is filled with water" stage: (stage 3)
      - <"sub-goal constraints", "the area of the handle of the teapot", "the normal of the handle of the teapot", "the area of the handle of the teapot rotates around the normal of the handle of the teapot by 30 degress">
      - <"flow constraints", "the cup is filled with water"> (go to stage 3 if satisfied; goto stage 4 if not satisfied)
    - "place the teapot on the table near the cup" stage: (stage 4)
      - <"sub-goal constraints", "the surface of the table", "the center of the cup opening of the cup", "the center of the teapot spout of the teapot is above the surface of the table by 10cm">
      - <"sub-goal constraints", "the center of the body of the teapot", "the center of the body of the cup", "the distance between the center of the body of the teapot and the center of the body of the cup is around 20cm">
      - <"sub-goal constraints", "the surface of the table", "the plane of the cup opening of the cup", "the surface of the table is parallel to the plane of the cup opening of the cup">

  - "put red block on top of the blue block":
    - "grasp red block" stage:
      - <"grasp", "the body of the red block">
      - <"sub-goal constraints", "the heading direction of the gripper approach of the robot", "the plane of the surface of the table", "the heading direction of the gripper approach of the robot is parallel to the normal of the surface of the table">
    - "drop the red block on top of blue block" stage:
      - <"sub-goal constraints", "the center of the red block", "the center of the blue block", "the center of the red block is directly above the center of the blue block around 20 centimeters">
    - "release the red block" stage:
      - <"release">
  - "open the door around the door hinge":
    - "grasp the door handle" stage:
      - <"grasp", "the handle of the door">
      - <"sub-goal constraints", "the heading direction of the gripper approach of the robot", "the plane of the door of the fridge", "the heading direction of the gripper approach of the robot is parallel to the normal of the door of the fridge">
      - <"sub-goal constraints", "the heading direction of the gripper binormal of the robot", "the heading direction of the handle of the fridge", "the heading direction of the gripper binormal of the robot is perpendicular to the heading direction of the handle of the fridge">
    - "rotate the door" stage:
      - <"sub-goal constraints", "the plane of the surface of the door", "the axis of the hinge of the door", "the plane of the surface of the door rotates around the axis of the hinge of the door by 90 degree">
      - <"path constaints", "the center of the handle of the door", "the axis of the hinge of the door", "the distance between the center of the handle of the robot and the hinge of the body of the door remains unchanged">
    - "release the door" stage:
      - <"release">
  - "cut the cucumber with the kitchen knife":
    - 'grasp the kitchen knife' stage:
      - <"grasp", "the handle of the kitchen knife">
      - <"sub-goal constraints", "the heading direction of the gripper approach of the robot", "the plane of the surface of the table", "the heading direction of the gripper approach of the robot is parallel to the normal of the surface of the table">
      - <"sub-goal constraints", "the heading direction of the gripper binormal of the robot", "the heading direction of the handle of the kitchen knife", "the heading direction of the gripper binormal of the robot is perpendicular to the heading direction of the handle of the kitchen knife">
    - "hang the knife above the cucumber"
      - <"sub-goal constaints", "the center of the blade of the kitchen knife", "the center of the body of the cucumber", "the center of the blade of the kitchen knife is directly above the center of the body of the cucumber by 20 cm">
      - <"sub-goal constaints", "the axis of the cucumber", "the plane of the blade of the knife", "the axis of the cucumber is perpendicular to the plane of the blade of the knife">
      - <"sub-goal constaints", "the heading direction of the blade of the knife", "the plane of the surface of the table", "the heading direction of the blade of the knife is parallel to the plane of the surface of the table">
    - "chop the cucumber" stage:
      - <"path constaints", "the axis of the cucumer", "the plane of the blade of the knife", "the axis of the cucumber remains perpendicular to the plane of the blade of the knife"> (remain from the previous constraints)
      - <"path constaints", "the heading direction of the blade of the knife", "the plane of the surface of the table", "the heading direction of the blade of the knife remains parallel to the plane of the surface of the table"> (remain from the previous constraints)
      - <"sub-goal constaints", "the center of the blade of the kitchen knife", "the center of the surface of the table", "the area of the blade of the kitchen knife is above the area of the surface of the table by 1 cm">
    - "release the cucumber" stage:
      - <"release">
  - "open the drawer":
    - "grasp the drawer handle" stage:
      - <"grasp", "the handle of the drawer">
      - <"sub-goal constraints", "the heading direction of the gripper of the robot", "the plane of the front face of the drawer", "the heading direction of the gripper of the robot is parallel to the normal of the front door of the drawer">
      - <"sub-goal constraints", "the heading direction of the gripper binormal of the robot", "the heading direction of the handle of the drawer", "the heading direction of the gripper binormal of the robot is perpendicular to the heading direction of the handle of the drawer">
    - "pull the drawer" stage:
      - <"sub-goal constraints", "the center of the handle of the drawer", "the center of the body the drawer", "the center of the handle of the drawer move backwards the center of the body of the drawer by around 30 cm">
    - "release the drawer" stage:
      - <"release">
  - "press the button"
    - "close the gripper" stage:
      - <"grasp", "">
    - "move to ready-to-press position" stage:
      - <"sub-goal constaints", "the heading direction of the robot approach of the robot", "the center of the body of the button", "the heading direction of the gripper approach of the robot colinear with the center of the body of the button">
      - <"path constaints", "the heading direction of the gripper of the robot", "the plane of the surface of the button", "the heading direction of the gripper of the robot remains parallel to the normal of the surface of the button">
    - "pressing" stage:
      - <"sub-goal constaints", "the center of the gripper of the robot", "the center of body of the button", "the center of the gripper of the robot reaches the center of the body of the button">
      - <"path constaints", "the heading direction of the gripper of the robot", "the plane of the surface of the button", "the heading direction of the gripper of the robot remains parallel to the normal of the surface of the button">
Example for geometric constraint generation and flow constraint generation under a single stage:
- Orbiting: We can only orbit each time 30 degrees due to design limitation. If we want to orbit for a circle, we need to repeatedly orbit 30 degrees by 12 times.
  - "orbit in one circle by x cm"
    - <"sub-goal constraints", "the center of A", "the center B", "the center of A orbit the center of B by 30 degrees">
    - <"path constraints", "the center of A", "the center of B", "the distance between the center of A and the center of B remains x cm">
    - <"flow constraints", "repeat this stage for 12 times (360 degrees in total)">
Flow constraint can be composed together to create complex flow constraint. Like loop-in-a-loop. Since there can be ONLY one flow constraint each stage, we need to have standalone stages to place the flow contraint like this:
Example:
- "..." (stage 3)
  - <"flow constraints", "condition"> (go to stage 3 if satisfied; goto stage 4 if not satisfied)
- "..." (stage 4)
  - <"flow constraints", "condition"> (go to stage 3 if satisfied; goto stage 5 if not satisfied)
\end{spverbatim}
\subsection{Prompts of Schemes for Cost Function Generation}
\begin{spverbatim}
    Please translate all the above geometric constraints and flow constaints for each stage into the Python cost function.
- We can obtain the point cloud by calling Python function "get_point_cloud(`the object part` of `the object`', `timestamp`)".
    - we record the position of the `geometry` since the grasping / contact stage, and record it into array.
    - specify `timestamp` to retrive `geometry` mask at the given timestamp. For example, timestamp = -2 to retrieve the previous mask at the time of grasping. timestamp = -1 to retrieve the current mask.
    - Example 1, if I want point cloud of "the axis of the body of the windmill" at its current timestamp, I can obtain the point cloud by  "mask = get_point_cloud('the body of the windmill', -1)". 
    - Example 2, if I want point cloud of "the plane of the surface of the door" at its previous timestamp, I can obtain the point cloud by "mask = get_point_cloud('the surface of the door', -2)".
- Please implement a Python cost function "stage_i_subgoal_constraints()", "stage_i_path_constraints()" for all the constraints tuples in the <> brackets one by one, except for the grasping and releasing constraints. It returns the cost measuring to what extent the constraint is satisfied. The constraint is satisfied when the cost goes down to 0. 
- Grasping, releasing should be a seperate sub-goal stage. 
- Implement "stage_i_flow_constraints()" for the flow constraint if needed, it returns the stage index to transit. If the flow constraints are not specified, we enter the next stage after this stage sequentially. Don't call undefined function in the flow constraint !
- For sub-goal constraint 1 of grasping , directly return grasp(`the object part` of `the object`). 
- You can specify multiple sub-goal constraints for grasping to specify the approach and binormal.
- For releasing in the sub-goal function directly return release().
- Constraint codes of each stage are splitted by a line "### <stage constraints splitter> ###"
- The unit of length is meter.
- The stage start from 1.
- Don't omit stages for the repeating stages, expand and list them one by one.
- Don't call function of other stage and is not defined, copy the function if necessary, but don't just call it.
- Left is -x axis, right is x axis, up is z axis, down is -z axis, front is y axis, back is -x axis.

Here are some examples:
### <stage constraints splitter> ### (if any)
### stage ? sub-goal constraints
def stage_?_subgoal_constraint1():
  """constraints: <"grasp", "the body of the banana"> """
    return grasp("the body of the banana")


### <stage constraints splitter> ###
### stage ? sub-goal constraints 
def stage_?_subgoal_constraint1():
    """constraints: <"sub-goal constraints", "the axis of the body of the cucumber", "the plane of the blade of the kitchen knife", "the axis of the body of the cucumber is perpendicular to the plane of the blade of the kitchen knife"> (for cutting cucumber)""" 
    pc1 = get_point_cloud("the body of the cucumber", -1)
    pc2 = get_point_cloud("the blade of the kitchen knife", -1)

    # Calculate the axis of the the body of the cucumber (pc1)
    # Compute the covariance matrix of the points in the point cloud
    covariance_matrix_cucumber = np.cov(pc1.T)
    # Get the eigenvalues and eigenvectors of the covariance matrix
    eigenvalues_cucumber, eigenvectors_cucumber = np.linalg.eig(covariance_matrix_cucumber)
    # The eigenvector corresponding to the largest eigenvalue is the axis of the body of the cucumber
    cucumber_axis = eigenvectors_cucumber[:, np.argmax(eigenvalues_cucumber)]
    if cucumber_axis[np.argmax(np.abs(cucumber_axis))] < 0:
      cucumber_axis = -cucumber_axis

    # Calculate the normal vector of the plane of the blade of the kitchen knife (pc2)
    covariance_matrix_knife = np.cov(pc2.T)
    eigenvalues_knife, eigenvectors_knife = np.linalg.eig(covariance_matrix_knife)
    # The eigenvector corresponding to the smallest eigenvalue is the normal vector of the surface
    knife_surface_normal = eigenvectors_knife[:, np.argmin(eigenvalues_knife)]
    if knife_surface_normal[np.argmax(np.abs(knife_surface_normal))] < 0:
      knife_surface_normal = -knife_surface_normal

    # Normalize both vectors
    cucumber_axis = cucumber_axis / np.linalg.norm(cucumber_axis)
    knife_surface_normal = knife_surface_normal / np.linalg.norm(knife_surface_normal)
    
    # Compute the dot product between the cucumber axis and knife surface normal
    dot_product = np.dot(cucumber_axis, knife_surface_normal)
    
    # cucumber_axis perpendicular to knife surface is to be parallel to the knife surface normal
    cost = (1 - abs(dot_product)) * 5.
    
    return cost

def stage_?_subgoal_constraint2():
    """constraints: <"sub-goal constraints", "the center of the body of the cucumber", "the center of the body of the kitchen knife", "the center of the body of the cucumber is directly above the center of the body of the kitchen knife by 10cm"> (for cutting cucumber)"""
    pc1 = get_point_cloud("the body of the cucumber", -1)
    pc2 = get_point_cloud("the body of the kitchen knife", -1)

    # Compute the mean position of the body the cucumber and the body of the kitchen knife
    body_of_cucumber_center = np.mean(pc1, axis=0)
    body_of_knife_center = np.mean(pc2, axis=0)
    
    # Calculate the horizontal distance (x, y coordinates) between the centers
    horizontal_distance = np.linalg.norm(body_of_cucumber_center[:2] - body_of_knife_center[:2])
    
    # Calculate the center of the body of the knife center should be 20 cm above the center of the body of the cucumber
    vertical_distance = body_of_knife_center[2] - 0.1 - body_of_cucumber_center[2]
    
    cost = abs(vertical_distance) + horizontal_distance
    
    return cost

def stage_?_subgoal_constraint3():
    """constraints: <"sub-goal constraints", "the heading direction of the blade of the knife", "the plane of the surface of the table", "the heading direction of the blade of the knife is parallel to the plane of the surface of the table"> (for cutting cucumber)""" 
    pc1 = get_point_cloud("the blade of the knife", -1)
    pc2 = get_point_cloud("the surface of the table", -1)

    # Calculate the heading direction vector of the plane of the blade of the knife (pc1)
    covariance_matrix_knife = np.cov(pc2.T)
    eigenvalues_knife, eigenvectors_knife = np.linalg.eig(covariance_matrix_knife)
    # The eigenvector corresponding to the smallest eigenvalue is the normal vector of the surface
    knife_surface_heading = eigenvectors_knife[:, np.argmin(eigenvalues_knife)]
    if knife_surface_heading[np.argmax(np.abs(knife
    _surface_heading))] < 0:
      knife_surface_heading = -knife_surface_heading

    # Calculate the normal vector of the plane of the surface of the table (pc2)
    covariance_matrix_table = np.cov(pc2.T)
    eigenvalues_table, eigenvectors_table = np.linalg.eig(covariance_matrix_table)
    # The eigenvector corresponding to the smallest eigenvalue is the normal vector of the surface
    table_surface_normal = eigenvectors_table[:, np.argmin(eigenvalues_table)]
    if table_surface_normal[np.argmax(np.abs(table_surface_normal))] < 0:
      table_surface_normal = -table_surface_normal

    # Normalize both vectors
    table_surface_normal = table_surface_normal / np.linalg.norm(table_surface_normal)
    knife_surface_heading = knife_surface_heading / np.linalg.norm(knife_surface_heading)
    
    # Compute the dot product between the table axis and knife surface normal
    dot_product = np.dot(table_surface_normal, knife_surface_heading)
    
    # knife surface heading parallel to the plane of the table surface is to be perpendicular to the table surface plane normal
    cost = abs(dot_product) * 5.
    return cost

def stage_?_subgoal_constraint1():
    """constraints: <"sub-goal constraints", "the plane of the surface of the door", "the axis of the hinge of the door", "the plane of the surface of the door rotate around the axis of the hinge of the door by 60 degrees"> (for opening the door)"""
    pc1 = get_point_cloud("the surface of the door", -1)
    pc1_previous = get_point_cloud("the surface of the door", -2)
    pc2 = get_point_cloud("the hinge of the door", -2)

    # Step 1: Normalize the axis of the hinge of the door (from pc2)
    covariance_matrix_door = np.cov(pc2.T)
    eigenvalues_door, eigenvectors_door = np.linalg.eig(covariance_matrix_door)
    door_axis = eigenvectors_door[:, np.argmax(eigenvalues_door)]
    door_axis = door_axis / np.linalg.norm(door_axis)  # Normalize the axis vector
    if door_axis[np.argmax(np.abs(door_axis))] < 0:
      door_axis= -door_axis

    # Step 2: Convert the angle from degrees to radians
    angle_radians = np.radians(angle_degrees)

    # Step 3: Compute the rotation matrix using Rodrigues' rotation formula
    K = np.array([[0, -door_axis[2], door_axis[1]],
                  [door_axis[2], 0, -door_axis[0]],
                  [-door_axis[1], door_axis[0], 0]])  # Skew-symmetric matrix for door_axis
    I = np.eye(3)  # Identity matrix
    rotation_matrix = I + np.sin(angle_radians) * K + (1 - np.cos(angle_radians)) * np.dot(K, K)

    # Step 4: Rotate each point in pc1
    rotated_pc1 = np.dot(pc1_previous - pc2.mean(0), rotation_matrix.T) + pc2.mean(0)  # Apply rotation matrix to each point

    # Step 5: compute the cost of how pc1 aligns with rotated_pc1.
    cost = np.linalg.norm(pc1 - rotated_pc1, axis=1).sum()
    return cost


### <stage constraints splitter> ###

### stage ? sub-goal constraints
def stage_?_subgoal_constraint1():
    """constraints: <"release"> """
    release()
    return

## Some geometry-related knowledge here:
{}
## End knowledge

Please write the codes below:
### <stage constraints splitter> ###
### stage 1 sub-goal constraints (if any)
## if it is a grasping constaints
def stage_1_subgoal_constraint1():
    """constraints: <"grasp", "`geometry` of `the object part' of `the object`"> """
    return grasp("`the object part' of `the object`")


def stage_1_subgoal_constraint1():
    """constraints: <?, ?, ?,..., ?>"""
    mask1 = get_point_cloud(?)
    mask2 = get_point_cloud(?)
    ...
    return cost
# Add more sub-goal constraints if needed
...

### stage 1 path constraints (if any)
def stage_1_path_constraint1():
    """constraints: <?, ?, ?, ?>"""
    mask1 = get_point_cloud(?)
    mask2 = get_point_cloud(?)
    ...
    return cost

# Add more path constraints if needed
...

Finally, write a list of "`geometry` of `the object part` of `the object`" in all the <> brackets:
object_to_segment = [?]
\end{spverbatim}
\subsection{Prompts of Geometric Knowledge for Cost Function Generation}
\begin{spverbatim}
Here are some geometry-related and control-flow-related knowledge:
THE EXAMPLES ARE ONLY FOR YOUR REFERENCE. YOU NEED TO ADAPT TO THE CODE FLEXIBLY AND CREATIVELY ACCORDING TO DIFFERENT SCENARIOS !

# Chapter 1: normal, axis, heading direction, binormal:
- Notice: The largest axis component of the normal / axis / heading direction should always be positive !
- To find the heading direction is the same of finding the axis
- Example:
    """
    Finds the normal (normal vector) of a plate given its point cloud.

    Args:
        pc: numpy array of shape (N, 3), point cloud of the plate.

    Returns:
        plate_normal: A normalized vector representing the normal vector of the plate.
    """
    # Compute the covariance matrix of the point cloud
    covariance_matrix = np.cov(pc.T)
    
    # Perform eigen decomposition to get eigenvalues and eigenvectors
    eigenvalues, eigenvectors = np.linalg.eig(covariance_matrix)
    
    # The eigenvector corresponding to the smallest eigenvalue is the normal vector to the plate's surface
    plate_normal = eigenvectors[:, np.argmin(eigenvalues)]
    if plate_normal[np.argmax(np.abs(plate_normal))] < 0:
        plate_normal = -plate_normal

    # Normalize the normal vector
    plate_normal = plate_normal / np.linalg.norm(plate_normal, axis=-1)
    
    return plate_normal

- Next example:
    """
    Finds the axis of a cylinder given its point cloud.

    Args:
        pc: numpy array of shape (N, 3), point cloud of the cylinder.

    Returns:
        cylinder_axis: A normalized vector representing the axis of the cylinder.
    """
    # Compute the covariance matrix of the point cloud
    covariance_matrix = np.cov(pc.T)
    
    # Perform eigen decomposition to get eigenvalues and eigenvectors
    eigenvalues, eigenvectors = np.linalg.eig(covariance_matrix)
    
    # The eigenvector corresponding to the largest eigenvalue represents the axis of the cylinder
    cylinder_axis = eigenvectors[:, np.argmax(eigenvalues)]
    if cylinder_axis[np.argmax(np.abs(cylinder_axis))] < 0:
        cylinder_axis = -cylinder_axis
    
    # Normalize the axis vector
    cylinder_axis = cylinder_axis / np.linalg.norm(cylinder_axis, axis=-1)
    
    return cylinder_axis
- To find out the heading direction of long-shaped object, find the max PCA component.
- To find out the normal of a surface, find the min PCA component.
- To find out the axis of an object, there are two cases. 
    - For long-shaped object like bolt, carrot, etc., its the max PCA component
    - For fat-shaped object like bowl, nut, etc., its the min PCA component

- A axis / heading direction / normal that is perpendicular to a plane / surface is parallel to the normal. 
- A binormal is the vector that is both perpendicular to the axis / heading direction and the normal 
- parallel: cost = (1 - np.abs(dot_product)) * 5

# Chapter 2: relative position between two points
- Example 1:
    """
    Measures the cost that point 2 is directly below point 1.
    
    Args:
        pc1: numpy array of shape (N, 3), point cloud of point 1.
        pc2: numpy array of shape (M, 3), point cloud of point 2.

    Returns:
        cost: a non-negative float representing the extent to which point 2 is directly below point 1.
              The lower the cost, the more point 2 is directly below point 1.
    """
    # Compute the center of mass (mean position) for point 1 and point 2
    point1_center = np.mean(pc1, axis=0)
    point2_center = np.mean(pc2, axis=0)
    
    # Calculate the horizontal distance (x, y coordinates) between the centers
    horizontal_distance = np.linalg.norm(point1_center[:2] - point2_center[:2])
    
    # Calculate the vertical distance (z coordinate) between the centers
    vertical_distance = point1_center[2] - point2_center[2]
    
    # If point 2 is not below point 1, add a large penalty to the cost
    if vertical_distance < 0:
        cost = abs(vertical_distance) + horizontal_distance + 1000  # Large penalty for incorrect vertical position
    else:
        cost = horizontal_distance
    
    return cost

- Next example:
    """
    Measures the cost that point 2 is directly to the left of point 1 by 10 cm.
    
    Args:
        pc1: numpy array of shape (N, 3), point cloud of point 1.
        pc2: numpy array of shape (M, 3), point cloud of point 2.

    Returns:
        cost: a non-negative float representing the extent to which point 2 is directly to the left of point 1 by 10 cm.
              The lower the cost, the closer point 2 is to being exactly 10 cm to the left of point 1.
    """
    # Compute the center of mass (mean position) for point 1 and point 2
    point1_center = np.mean(pc1, axis=0)
    point2_center = np.mean(pc2, axis=0)
    
    # Calculate the horizontal distance (x-axis) between point 1 and point 2
    x_distance = point2_center[0] - point1_center[0]
    
    # Calculate the y and z distances (vertical and depth positions)
    y_distance = abs(point2_center[1] - point1_center[1])
    z_distance = abs(point2_center[2] - point1_center[2])
    
    # The ideal x distance should be -0.10 meters (to the left by 10 cm)
    cost = abs(x_distance + 0.10) + y_distance + z_distance  # Sum all deviations from ideal positioning
    
    return cost

# Chapter 3: control flow
We use flow constraints for control flow, which specify transitions among different stages.
- Repetition control flow: Do <something> until some <condition>
- For example:
<"flow constraint", "Repeat this stage until the box reaches the table edge">
def stage_`i`_flow_constraint1():
  while True:
    # query GPT-4O
    query = "Is the box on the table edge? You only need to answer 'yes' or 'no'"
    answer = query_GPT(query)
    if answer.strip().lower() == "yes"
      return `i+1` # go to next stage 
    else:
      return `i` # repeat this stage to continue pushing the box
## Repeat until the cup is being filled, then go to stage 3
## <"flow constraints", "the cup is filled with water">
def stage_i_flow_constraint1():
  while True:
    # query GPT-4O
    query = "Is the water filled in the cup? You only need to answer 'yes' or 'no'"
    answer = query_GPT(query)
    if answer.strip().lower() == "yes"
      return `i+1`
    else:
      return `i`

## Repeat the stage N times
## <"flow constraints", "repeat this stage N times">
def stage_i_flow_constraint1():
  # CNT is a global counter variable with default value 0, don't initialize it again!
  if CNT < N:
    CNT += 1
    return `i`
  CNT = 0
  return `i+1`
- You can have multiple flow constraint if necessary. They can create complex flow control. Just think about what you do to write flow control in Python code.
For a example:
## <"flow constraints", "repeat this stage N times">
## <"flow constraints", "condition">
This is example of loop in a loop. The inner loop repeat the stage N times. The outer loop repeat the inner loop until condition is satisfied.
Another example:
## <"flow constraints", "repeat this stage N times">
## <"flow constraints", "condition">

# Chapter 4: rotation and orbiting
- To rotate, we use sub-goal constraint to first constraints its rotated position
## rotate pc around axis by angle_degrees
def stage_?_subgoal_constraint1():
    pc_previous = get_point_cloud("pc", -2)
    pc = get_point_cloud("pc", -1)
    object = get_point_cloud("object", -2) # use -2 to specify the previous object
    covariance_matrix = np.cov(object.T)
    eigenvalues, eigenvectors = np.linalg.eig(covariance_matrix)
    axis = eigenvectors[:, np.argmax(eigenvalues)]
    axis = axis / np.linalg.norm(axis, axis=-1)  # Normalize the axis vector

    # Step 3: Convert the angle from degrees to radians
    angle_radians = np.radians(angle_degrees)

    # Step 4: Compute the rotation matrix using Rodrigues' rotation formula
    K = np.array([[0, -axis[2], axis[1]],
                  [axis[2], 0, -axis[0]],
                  [-axis[1], axis[0], 0]])  
    I = np.eye(3)  # Identity matrix
    rotation_matrix = I + np.sin(angle_radians) * K + (1 - np.cos(angle_radians)) * np.dot(K, K)

    # Step 5: Rotate each point in pc1 around object's center
    rotated_pc = np.dot(pc_previous - object.mean(0), rotation_matrix.T) + object.mean(0)

    cost = np.linalg.norm(rotated_pc - pc, axis=-1).sum()
    return cost

- To orbit: The orientation of pc is unchanged during orbiting. To calculate the position after orbital translation, we first calculate the position of the center of pc rotating around the axis of the object. Next, we translate the whole pc to the rotated center.
def stage_?_subgoal_constraint1():
    pc_previous = get_point_cloud("pc", -2)
    pc = get_point_cloud("pc", -1)
    object = get_point_cloud("object", -2) # use -2 to specify the previous object
    covariance_matrix = np.cov(object.T)
    eigenvalues, eigenvectors = np.linalg.eig(covariance_matrix)
    axis = eigenvectors[:, np.argmax(eigenvalues)]
    axis = axis / np.linalg.norm(axis, axis=-1)  # Normalize the axis vector
    # Step 3: Convert the angle from degrees to radians
    # Step 4: Compute the rotation matrix using Rodrigues' rotation formula
    # Step 5: Rotate each point in pc1 around object's center
    orbital_pc_center = np.dot(pc_previous.mean(0) - object.mean(0), rotation_matrix.T) + object.mean(0)
    orbital_pc = orbital_pc - pc_previous.mean(0) + orbital_pc_center
    cost = np.linalg.norm(rotated_pc - pc, axis=-1).sum()
    return cost

- For both rotation and orbiting, if the distance is not specified, we need a path constraint to specify the distance between pc center and the object center remain unchanged (same as the distance of pc_previous center and the object center)
def stage_?_path_constraint1():
     pc_previous = get_point_cloud("pc", -2)
    pc = get_point_cloud("pc", -1)
    object = get_point_cloud("object", -2) # use -2 to specify the previous object
    distance_previous = np.linalg.norm(pc_previous.mean(0) - object.mean(0))
    distance = np.linalg.norm(pc.mean(0) - object.mean(0))
    cost = abs(distance_previous - distance)
    return cost
- If certain distance `x` is specified, we need path constraint to remain the specified distance:
def stage_?_path_constraint1():
    # get pc, and object
    distance = np.linalg.norm(pc.mean(0) - object.mean(0))
    cost = abs(distance - x)
    return cost
- To turn something, rotate all its points around its axis by some angle.
- To orbit in circle, using flow control to repeat this stage 12 times: <"flow constraints", "repeat this stage 12 times">. For sub-goal constraint, orbit by 30 angle_degrees.
- To rotate / orbit clockwisely, the angle is negative; Otherwise, the angle is positive.


# Chapter 4: Relationship between points and vector
- Colinear: point B colinear with object A's axis / normal / heading direction by distance x if:
    point B = point A's center + normalize(point A's axis / normal / heading direction) * x
- move towards / backwards / against / away:
    - We need to calculate the target point first and calculate the distance between previous point and the target point as the cost
    - points A move towards / to points B by distance:
        previous point A = get_point_cloud(A, -2)
        current point A = get_point_cloud(A, -1)
        moving direction = normalized(vector of previous point A to B)
        target position of point A = points A  + moving direction * distance
        cost = np.linalg.norm(target position of point A - current position of point A) ## the cost is calculated based on the distance between target point and current point !!
    - points A move backward / against / away from points B by distance:
        previous point A = get_point_cloud(A, -2)
        current point A = get_point_cloud(A, -1)
        moving direction = normalized(vector of previous point A to B)
        target position of point A = points A + moving direction * distance
        cost = np.linalg.norm(target position of point A - current position of point A) ## the cost is calculated based on the distance between target point and current point

\end{spverbatim}
\subsection{Prompts of Schemes for Segmentation Mask Selection and Processing}
\begin{spverbatim}
    There are totally {number of pair} pair of images. 
For each pair, the left image is the image of {object name} with different part highlighted in red. The right image is the segmentation mask highlighted in white to represent different parts of {object name}. These images are named as image i, ... (i=0, 1, 2, ...)
Please infer what is highlighted in red for the left image one by one, and then select one of the image for {geometric part name}.
- Output: image {image_index}, `geometry` (i=0,1,2... is the index number) at the end in a single line.
- Where `geometry` is the geometry of object, like the edge, the center, the area, left point, right, point, etc..

Write a Python function to find out the {geometric part name} given the segmentation of image {object name}, {image_index}. 
- the input `mask` is a boolean numpy array of a segmentation mask in shapes (H, W)
- return the mask which is a numpy array. 
- You can `import numpy as np` and `import cv2`, but don't import other packages
- mask_output should still be in the shape(H, W)
## code start here
def segment_object(mask):
    ...
    return mask_output
Please directly output the code without explanations. Complete the comment in the code. Remove import lines since they will be manually imported later.
\end{spverbatim}
\subsection{Prompts of Segmentation Knowledge for Segmentation Mask Selection and Processing}
\begin{spverbatim}
    - To find hinge / axis, output the image of its door, and see which side to segment.For a rotating object part, the hinge / axis and the handle are of the opposite position.  For example, for finding the hinge of the microwave, output the image of microwave door first. And if the handle is on the left of the the door, the hinge should locate at the right edge of its door. 
- For a sliding body, the slider should be parallel to the edge of the frame.
- sample code to find the complete edge. You need to adjust the code to choose the left / right / top / bottom edge accordingly. For example, to fine the left edge, find the leftmost True value by iterating over each row to find the leftmost True value
    
def find_edges(mask):
     """
    Find the edges of a binary mask using Canny edge detection.

    Parameters:
        mask (np.ndarray): Binary image (mask) with 1s representing the object and 0s representing the background.
        
    Returns:
        np.ndarray: Edge mask with 255 at the edges of the object and 0s elsewhere.
    """
    # Convert mask to uint8 if not already
    mask = (mask * 255).astype(np.uint8) if mask.max() == 1 else mask
    
    # Apply Canny edge detection
    edges = cv2.Canny(mask, 100, 200)
    
    # shift the edge down a little bit !
    edges = np.roll(edges, 3, axis=0)

    # Set the top rows to zero to prevent wrap-around artifacts
    edges[:3, :] = 0

    return edges

- return the mask directly if the mask does not need to be processed
\end{spverbatim}

\section{MetaWorld Environment}
\subsection{Task Descriptions}
We design 6 tasks for the MetaWorld environment: 
\begin{itemize}
    \item Btn-press. Task description: press the red button from its side. Success condition: The red button is entirely pressed.
    \item Btn-press-top. Task description: Press the red button from top-down. Success condition: The red button is pressed entirely.
    \item Handle-press. Task description: Press the red handle. Success condition: The handle is entirely pressed.
    \item Shelf-place. Task description: Put the blue cube onto the middle stack of the shelf. Grasp the blue cube and lift it vertically before moving to the middle stack of the shelf. Success condition: The blue cube is on the middle stack of the shelf.
    \item Basketball. Task description: Put the basketball onto the hoop. Lift the ball vertically and move over above the hoop, Success condition: The basketball pass through the hoop.
    \item Assembly. Task description: Put the round ring into the red stick. Grasp the green handle of the round ring and put the hole into the red stick. Success condition: The red stick is inside the round ring.
\end{itemize}
\subsection{Manipulation Visualization}
\begin{figure}[H]
    \centering
    \includegraphics[width=\linewidth]{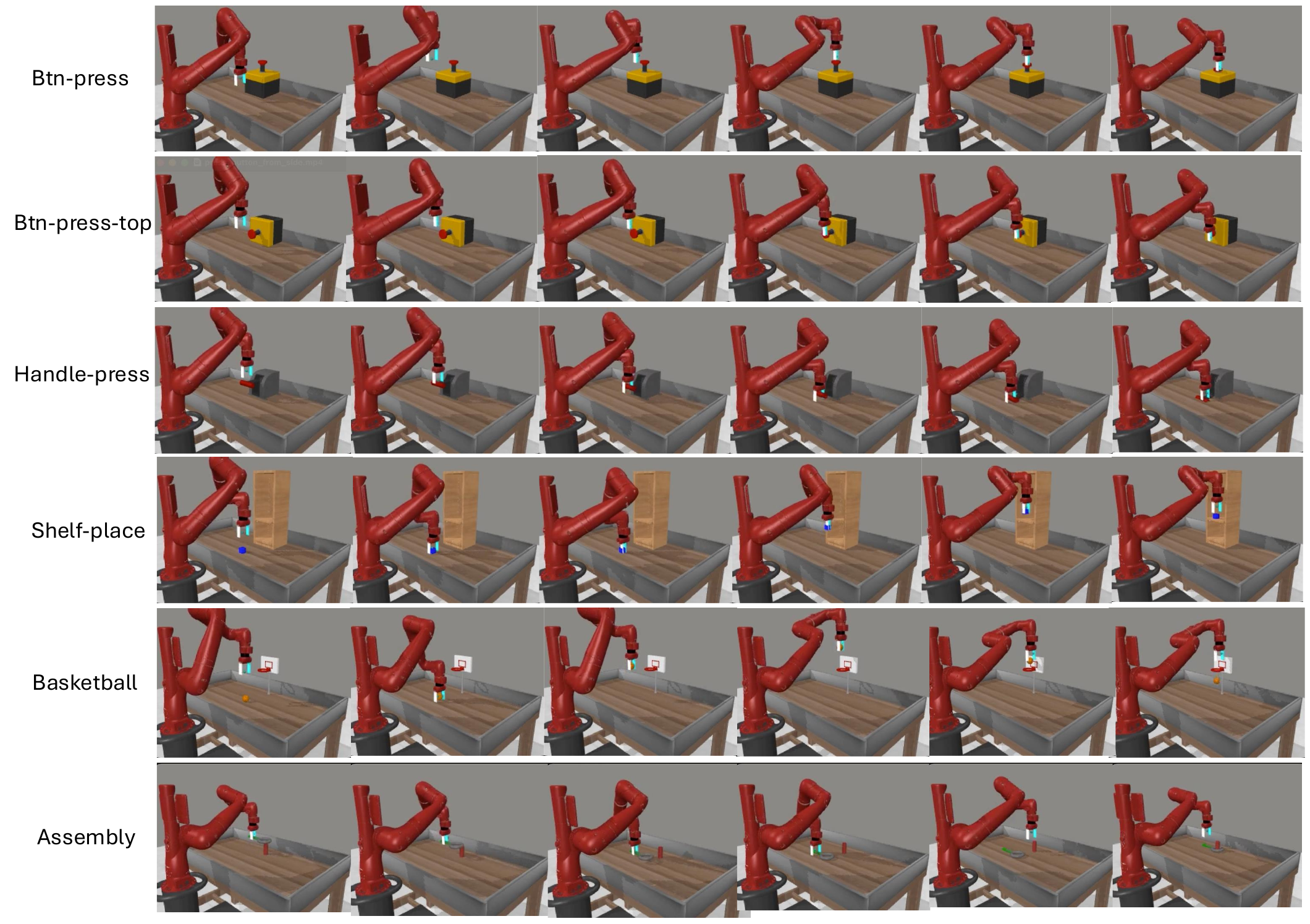}
    \caption{Visualization of execution for each task in MetaWorld environment.}
    \label{fig:enter-label}
\end{figure}
\section{Omnigibson Environment}
\subsection{Task Descriptions}
For Omnigibson environment, we design 4 tasks:
\begin{itemize}
    \item Cut-carrot. Task description: Cut the carrot with the knife. Success condition: The knife blade intersect with the carrot top-down with its normal perpendicular to the carrot's heading direction.
    \item Open-fridge. Task description: Open the fridge. Success condition: The fridge door is open by at least 45 degrees.
    \item Put-pen-into-holder. Task description: Put the pen perpendicularly into the black cup. Success condition: The pen is inside the pen holder.
    \item Typing. Task description: Type "hi" on the computer keyboard. Success condition: The "H" key and the "I" key on the computer keyboard are pressed sequentially.
\end{itemize}
\subsection{Manipulation Visualization}
\begin{figure}[H]
    \centering
    \includegraphics[width=\linewidth]{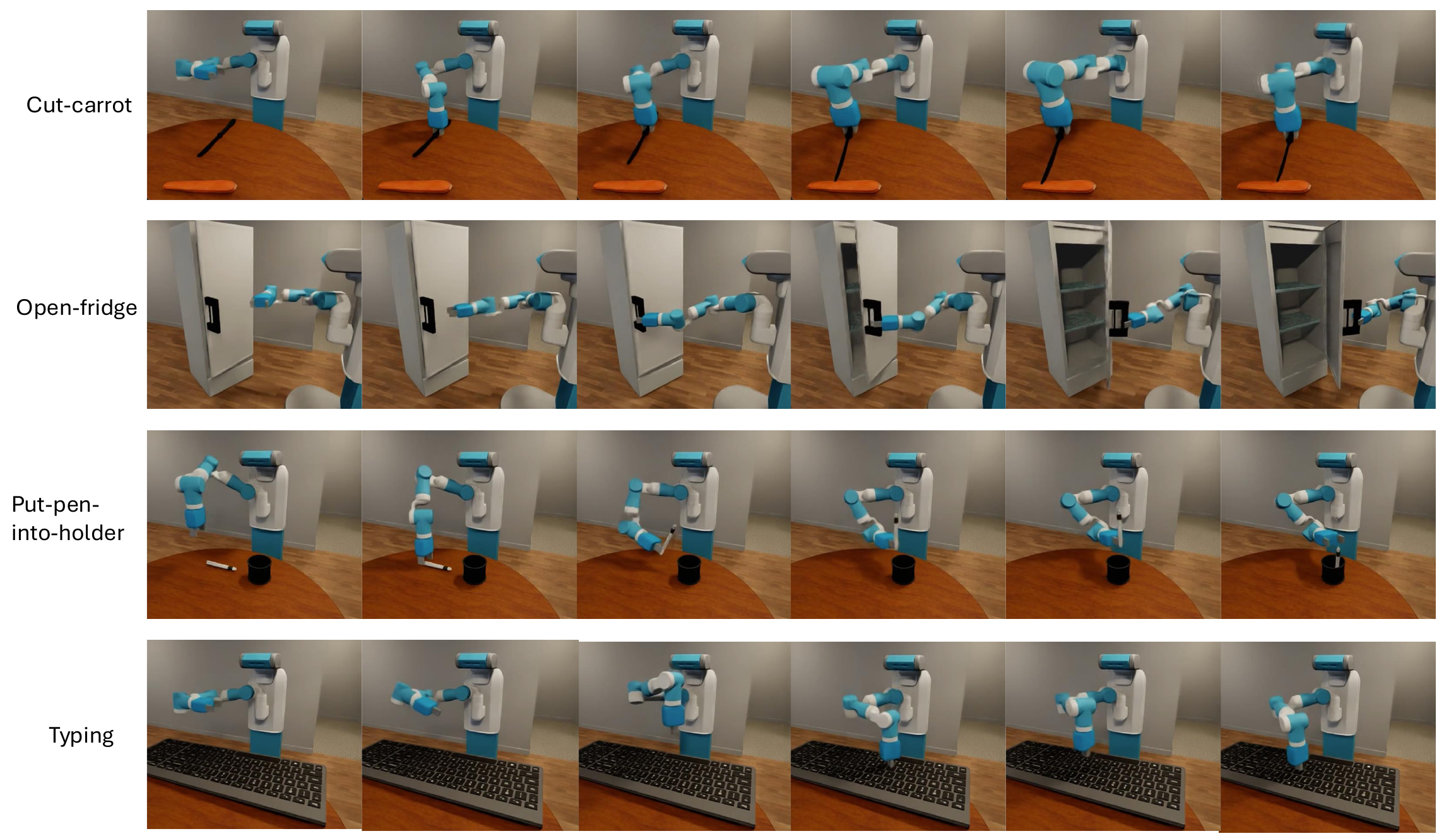}
    \caption{Visualization of execution for each task in Omnigibson environment.}
    \label{fig:enter-label}
\end{figure}
\section{Real World Environment}
\subsection{Task Descriptions}
We design 4 tasks for the real-world environment:
\begin{itemize}
    \item Pick-place. Task description: Put $<$object A$>$ into / onto $<$object B$>$. Success condition: $<$object A$>$ is inside / on $<$object B$>$.
    \item Pour: Task description: Fill $<$object A$>$ with $<$object B$>$. Success condition: $<$object B$>$ is filled with some $<$object A$>$
    \item Open: Task description: open $<$object$>$. Success condition: $<$object$>$ is open by at least 30 degrees / 5 cm.
    \item Stir. Task description: Stir $<$object A$>$ with $<$object B$>$. Success condition: $<$object B$>$ moves periodically inside $<$object A$>$.
\end{itemize}

\end{document}